%% file: icml2023.tex
\documentclass{article}

\usepackage{microtype}
\usepackage{graphicx}
\usepackage{subfigure}
\usepackage{multirow}
\usepackage{booktabs} % for professional tables
\usepackage{threeparttable}
\usepackage{hyperref}

\usepackage[arxiv]{icml2023}

\usepackage{amsmath}
\usepackage{amssymb}
\usepackage{mathtools}
\usepackage{amsthm}
\usepackage{wrapfig}
\usepackage[capitalize,noabbrev]{cleveref}

\theoremstyle{plain}

\theoremstyle{definition}

\theoremstyle{remark}

\usepackage{bm}
\usepackage{mathtools}
\usepackage[textsize=tiny]{todonotes}

\DeclareMathOperator*{\argmax}{arg\,max}

\icmltitlerunning{Cut your Losses with Squentropy}

\begin{document}

\twocolumn[
\icmltitle{Cut your Losses with Squentropy}



\icmlsetsymbol{equal}{*}

\begin{icmlauthorlist}
\icmlauthor{Like Hui}{lh,mb}
\icmlauthor{Mikhail Belkin}{mb,lh}
\icmlauthor{Stephen Wright}{sw}

% Hal{\i}c{\i}o{\u g}lu Data Science Institute\\
% University of California,s San Diego

\end{icmlauthorlist}

\icmlaffiliation{lh}{Computer Science and Engineering, University of California, San Diego}
\icmlaffiliation{mb}{Halıcıoğlu Data Science Institute, University of California, San Diego}
\icmlaffiliation{sw}{Wisconsin Institute for Discovery, UW-Madison}

\icmlcorrespondingauthor{Like Hui}{lhui@ucsd.edu}
\icmlcorrespondingauthor{Mikhail Belkin}{mbelkin@ucsd.edu}
\icmlcorrespondingauthor{Stephen Wright}{swright@cs.wisc.edu}

\icmlkeywords{Machine Learning, ICML}

\vskip 0.3in
]


\printAffiliationsAndNotice{}  % leave blank if no need to mention equal contribution

\begin{abstract}
Nearly all practical neural models for classification are trained using cross-entropy loss. Yet this ubiquitous choice is supported by little theoretical or empirical evidence. 
Recent work \cite{hui2020evaluation} suggests that training using the (rescaled) square loss is often superior in terms of the classification accuracy. 
In this paper we propose the ``squentropy'' loss, which is the sum of two terms:  the cross-entropy loss and  the average square loss over the incorrect classes. 
We provide an extensive set of experiments on multi-class classification problems showing that the squentropy loss outperforms both the pure cross entropy and rescaled square losses in terms of the classification accuracy. 
We also demonstrate that it provides significantly better model calibration than either of these alternative losses and, furthermore, has less variance with respect to the random initialization. 
Additionally, in contrast to the square loss, squentropy loss can typically  be trained using exactly the same optimization parameters, including the learning rate, as the standard cross-entropy loss, making it a true ``plug-and-play'' replacement.   Finally, unlike the rescaled square loss, multiclass squentropy contains no parameters that need to be adjusted.

\end{abstract}

\section{Introduction}

As with the choice of an optimization algorithm, the choice of loss function is an indispensable ingredient in training neural network models. Yet, while there is extensive theoretical and empirical research into optimization and regularization methods for training deep neural networks~\cite{sun2019optimization}, far less is known about the selection of loss functions.  
 In recent years, cross-entropy loss has been predominant in training for multi-class classification with modern neural architectures. 
 There is surprisingly little theoretical or empirical evidence in support of this choice.  
 To the contrary, an extensive set of experiments with neural architectures  conducted in~\cite{hui2020evaluation} indicated that training with the (rescaled) square loss produces similar or  better classification accuracy than cross entropy on most classification tasks. 
 Still, the rescaled square loss proposed  in  that work requires additional parameters (which must be tuned) when the number of classes is large. 
 Further, the optimization learning rate for the square loss is typically different from that of cross entropy, which precludes the use of square loss as an out-of-the-box replacement. 

In this work we propose the ``squentropy'' loss function for multi-class classification. Squentropy is the sum of two terms: the standard cross-entropy loss and the average square loss over the incorrect classes. Unlike the rescaled square loss, squentropy has no adjustable parameters. Moreover, in most cases, we can simply use the optimal hyperparameters for cross-entropy loss without any additional tuning, making it a true  ``plug-and-play'' replacement for cross-entropy loss.

To show the effectiveness of squentropy, we provide comprehensive experimental results over a broad range of benchmarks with different neural architectures and data from NLP, speech, and computer vision.  In {24 out of 34} tasks, squentropy has the best (or tied for best) classification accuracy, in comparison with cross entropy and the rescaled square loss. 
Furthermore, squentropy has consistently improved {\it calibration}, an important measure of how the output values of the neural network match the underlying probability of the labels~\cite{guo2017calibration}. 
Specifically, in  { 26 out of 32} tasks for which calibration results can be computed, squentropy is better calibrated than either alternative. 
We also show results on 121 tabular datasets from~\cite{fernandez2014we}. Compared with cross entropy, squentropy has better test accuracy on {94 out of 121} tasks, and better calibration on 83 datasets.  
Finally, we show that squentropy is less sensitive to the randomness of the initialization than either of the two alternative losses.

Our empirical evidence suggests that in most settings, squentropy should be the first choice of loss function for multi-class classification via neural networks.

\section{The squentropy loss function}
\label{sec_squentropy}

The problem we consider here is supervised multi-class classification. 
We focus on the loss functions for training neural classifiers on this task.

Let $D=(\bm{x}_i, y_i)_{i=1}^n$ denote the dataset sampled from a joint distribution $\mathcal{D}(\mathcal{X}, \mathcal{Y})$. For each sample $i$, $\bm{x}_i\in \mathcal{X}$ is the input and $y_i \in \mathcal{Y}=\{1,2, \dotsc, C\}$ is the true class label. The one-hot encoding label used for training is $\bm{e}_{y_i}=[0,\ldots, \underbrace{1}_{y_i}, 0, \ldots,0]^T \in \mathbb{R}^C$.
Let $f(\bm{x}_i)\in \mathbb{R}^C$ denote the logits (output of last linear layer) of a neural network of input $\bm{x}_i$, with components $f_j(\bm{x}_i)$, $j=1,2,\dotsc,C$.
Let $p_{i,j}=e^{f_j(\bm{x}_i)}/\sum_{j=1}^C e^{f_j(\bm{x}_i)}$ denote the predicted probability of $\bm{x}_i$ to be in class $j$. 
Then the squentropy loss function on a single sample $\bm{x}_i$ is defined as follows:
\begin{equation}
\label{mix_func}
    l_{\text{sqen}}(\bm{x}_i,y_i) = - \log p_{i,y_i}(\bm{x}_i) + \frac{1}{C-1}\sum_{\substack{j=1,}{j\neq y_i}}^C f_j(\bm{x}_i)^2.
\end{equation}
The first term  $- \log p_{i,y_i}(\bm{x}_i)$ is simply cross-entropy loss. 
The second term is the square loss averaged over the incorrect ($j\neq y_i$) classes. 

The cross-entropy loss is minimized when $f_{y_i}(\bm{x}_i) \to \infty$ while $f_j(\bm{x}_i) \to - \infty$ or at least stays finite for $j \ne y_i$. By encouraging all incorrect logits  to go to a specific point, namely $0$, it is possible that squentropy yields a more ``stable'' set of logits ---  the potential for the incorrect logits to behave chaotically is taken away. In other words, the square loss term plays the role of a regularizer. 
We discuss this point further in Section~\ref{sec_norm}.
\paragraph{Dissecting squentropy.}  Cross entropy  acts as an effective penalty on the prediction error made for  the true class $y_i$, as it has high loss and large gradient when $p_{i,y_i}$ is close to zero, leading to effective steps in a gradient-based  optimization scheme. 
% \lhcomment{which one is the correct reference?}
The ``signal'' coming from the gradient for the incorrect classes is weaker, so such optimization schemes may be less effective in driving the probabilities for these classes to zero.
Squentropy can be viewed as a modification of the rescaled square loss~\cite{hui2020evaluation}, in which cross entropy replaces the term $t(f_{y_i}(\bm{x}_i)-M)^2$ corresponding to the true class, which depends on two parameters $t$, $M$ that must be tuned. 
This use of cross entropy dispenses with the additional parameters yet provides an adequate ``signal'' for the gradient for a term that captures loss on the ``true'' class.

The second term in \eqref{mix_func} pushes all logits $f_j(\bm{x}_i)$ corresponds to false classes $j \ne y_i$ to $0$.
Cross entropy attains a loss close to zero on term $i$ by sending $f_{y_i}(\bm{x}_i)\to \infty$ and/or $f_j(\bm{x}_i) \to -\infty$ for all $j \neq y_i$. 
By contrast, squentropy ``anchors" the incorrect logits at zero (via the second term) while driving $f_{y_i}(\bm{x}_i)\to \infty$  (via the first term). Then the predicted probability of true class $p_{i, y_i}(\bm{x}_i)$ will be close to $\frac{e^{f_{y_i}(\bm{x}_i)}}{e^{f_{y_i}(\bm{x}_i)}+C-1}$ for squentropy, which possibly approaches $1$ more slowly than for cross entropy. 
When the training process is terminated, the probabilities $p_{i, y_i}(\bm{x}_i)$ tend to be less clustered near $1$ for squentropy than for cross entropy. 
Confidence in the true class thus tends to be slightly lower in squentropy.
We see the same tendency toward lower confidence in the  {\em test} data, thus helping calibration. 

In calibration literature, various post-processing methods, such as Platt scaling \cite{platt1999probabilistic} and temperature scaling \cite{guo2017calibration}, also improves calibration by reducing $p_{i,y_i}$ below $1$, while other methods such as label smoothing \cite{muller2019does, liu2022devil} and  focal loss \cite{mukhoti2020calibrating} achieve similar reduction on the predicted probability.
While all these methods require additional hyperparameters, squentropy does not. We conjecture that calibration of squentropy can be further improved by combining it with these techniques. 
\paragraph{Relationship to neural collapse.}
Another line of work that motivates our choice of loss function is the concept of neural collapse \cite{Pap20a}. 
Results and observations for neural collapse interpose a linear transformation between the outputs of the network (the transformed features $f_j(\bm{x}_i)$) and the loss function. They show broadly that the features collapse to a class average and that, under a cross-entropy loss, the final linear transformation maps them to rays that point in the direction of the corners of the simplex in $\mathbb{R}^C$.
(A modified version of this claim is proved for square loss in \cite{han2021neural}.)
Our model is missing the interposing linear transformation, but these observations suggest roughly that cross entropy should drive the true logits $f_{y_i}(\bm{x}_i)$ to $\infty$ while the incorrect logits $f_j(\bm{x}_i)$ for $j \ne y_i$ tend to drift toward $-\infty$, as discussed above.
As noted earlier, the square loss term in our squentropy loss function encourages $f_j(\bm{x}_i)$ for $j \ne y_i$ to be driven to zero instead --- a more well defined limit and one that may be achieved without blowing up the weights in the neural network (or by increasing them at a slower rate). 
In this sense, as mentioned above, the squared loss term is a kind of regularizer.
\paragraph{Confidence calibration.} We use the expected calibration error (ECE) \cite{naeini2015obtaining} to evaluate confidence calibration performance.
It is defined as $\mathbb{E}_{p}[|\mathbb{P}(\hat{y}=y|p)-p|]$, where $p$ and $y$ correspond to the estimated probability and true label of a test sample $\bm{x}$. $\hat{y}$ is the predicted label given by $\argmax_j p_j$. It captures the expected difference between the accuracy $\mathbb{P}(\hat{y}=y|p)$ and the estimated model confidence $p$. 
Because we only have finite samples in practice, and because we do not have access to the true confidences $p_{\text{true}}$  for the test set (only the labels $y)$, we need to replace this definition with an {\em approximate} ECE. 
This quantity is calculated by dividing  the interval $[0,1]$ of probability predictions into $K$ equally-spaced bins with the $k$-th bin interval to be $(\frac{k-1}{K}, \frac{k}{K}]$. 
Let $B_k$ denote the set of test samples $(\bm{x}_i,\hat{y}_i)$ for which the confidence  $p_{i,y_i}$ predicted by the model lies in bin $k$. 
(The probabilities $p_{i,j}$ are obtained from a softmax on the exponentials of the logits $f_j(\bm{x}_i)$.)
The accuracy of this bin is defined to be $\text{acc}(B_k)=\frac{1}{|B_k|}\sum_{i\in B_k}\mathbf{1}(\hat{y}_i=y_i)$, where $y_i$ is the true label for the test sample $\bm{x}_i$ and $\hat{y}_i$ is the model prediction for this item (the one for which $p_{i,j}$ are maximized over $j=1,2,\dotsc,C$).
The confidence for bin $k$ is defined empirically as  $\text{conf}(B_k) = \frac{1}{|B_k|}\sum_{i\in B_k} p_{i,y_i}$. 
We then use the following definition of ECE:
\begin{equation} \label{eq:ece}
    \text{ECE} = \sum_{k=1}^K\frac{|B_k|}{n}\left|\text{acc}(B_k)-\text{conf}(B_k)\right|.
\end{equation}
This quantity is small when the frequency of correct classification over the test set matches the probability of the predicted label.

\section{Experiments}
In this paper we  consider three loss functions, our proposed squentropy, cross entropy and the (rescaled) square loss from \cite{hui2020evaluation}. The latter is formulated as follows:
\begin{equation}
\label{square_func}
    {l}_{s}(\bm{x}_i,y_i) = \frac{1}{C}\left(t*(f_{y_i}(\bm{x}_i)-M)^2+\sum_{j=1, j\neq y_i}^Cf_j(\bm{x}_i)^2\right),
\end{equation}

where $t$ and $M$ are positive parameters.
($t=M=1$ yields standard square loss.)
We will point out those entries in which values  $t>1$ or $M>1$ were used; for the others, we set $t=M=1$. 
Note that following \cite{hui2020evaluation}, the square loss is directly applied to the logits, with no softmax layer in training.

We conduct extensive experiments on various datasets.
These include a wide range of well-known benchmarks across NLP, speech, and vision with different neural architectures ---  more than $30$ tasks altogether. 
In addition, we evaluate the loss functions on 121 tabular datasets \cite{fernandez2014we}. 
In the majority of our experiments, training with squentropy gives best test performance and  also consistently better calibration results. 

\paragraph{Training scheme.} In most of experiments we train with squentropy with hyperparameter settings that are optimal for cross entropy, given in \cite{hui2020evaluation}. This choice favors cross entropy.
% , as those hyperparameters have been tuned for optimal performance for cross entropy. 
This choice also means that switching to squentropy  requires a change of just one line of code.
Additional gains in performance of squentropy might result from additional tuning, at the cost of more computation in the hyperparameter tuning process.

\input{accuracy_ece}

\paragraph{Datasets.} We test on a wide range of well-known benchmarks from NLP, speech and computer vision. NLP datasets include MRPC, SST-2, QNLI, QQP, text8, enwik8, text5, and text20. 
Speech datasets include TIMIT, WSJ, and Librispeech. 
MNIST, CIFAR-10, STL-10 \cite{coates2011analysis}, CIFAR-100, SVHN \cite{netzer2011reading}, and ImageNet are vision tasks. See Appendix A of \cite{hui2020evaluation} for details of most of those datasets.
(The exceptions are SVHN, STL-10, and CIFAR-100, which we describe in Appendix \ref{app_data} of this paper). 
The 121 tabular datasets are from \cite{fernandez2014we} and they are mostly small datasets --- $90$ of them have $\le 5000$ samples. The feature dimension is small (mostly $<50$) and most datasets are class-imbalanced. 
\vspace{-1mm}
\paragraph{Architectures and hyperparameter settings.} We choose various modern neural architectures, including simple fully-connected networks, convolutional networks (TCNN\cite{bai2018empirical}, Resnet-18, VGG, Resnet-50 \cite{he2016deep}, EfficientNet\cite{tan2019efficientnet}), LSTM-based networks \cite{chen2016enhanced} (LSTM+CNN, LSTM+Attention, BLSTM), and Transformers \cite{vaswani2017attention} (fine-tuned BERT, Transformer-XL, Transformer, Visual transformer). 
See Table \ref{acc_ece} for detailed references. 
We follow the hyperparameter settings given in Appendix~\ref{app_para} of \cite{hui2020evaluation} for the cross-entropy loss and the square loss (other than SVHN, STL-10, and CIFAR-100), and use the algorithmic parameters settings of the cross entropy for squentropy in most cases.
The exceptions are SVHN and STL-10, where squentropy and square loss have a smaller learning rate (0.1 for cross entropy while 0.02 for squentropy and square loss). 
More details about hyperparameter settings of SVHN, STL-10, CIFAR-100 are in Appendix \ref{app_para}.

\paragraph{Metrics.} For NLP, vision and 121 tabular datasets, we report accuracy as the metric for test performance. For speech dataset, we conduct the automatic speech recognition (ASR) tasks and report test set error rates which are standard metrics for ASR. Precisely, for TIMIT, we report phone error rate (PER) and character error rate (CER). For WSJ and Librispeech, we report CER and word error rate (WER). ECE is the metric to measure the calibration results for all datasets. For speech datasets, we report calibration results for the acoustic modeling part. See Table \ref{acc_ece} shows the results of NLP, speech and vision datasets. Figure \ref{121_results} show results of 121 tabular datasets. In addition, we provide reliability diagrams \cite{degroot1983comparison, niculescu2005predicting} to visualize the confidence and accuracy of each interval and see details in Section \ref{calibration_expr}.

\paragraph{Remarks on Table \ref{acc_ece}.} For the results of square loss, we use rescaled square loss with $t>1$ or $M>1$ for TIMIT(PER) ($t=1, M=15$), WSJ ($t=1, M=15$), Librispeech ($t=15, M=30$), CIFAR-10 and CIFAR-100 ($t=1, M=10$), and ImageNet ($t=15, M=30$). All others are the standard square loss. Note that WSJ (WER) and WSJ (CER) share the same ECE number as they share one acoustic model. (Similarly for Librispeech.)
Additionally, since ECE numbers are not available for Top-5 accuracy, the corresponding entries  (ImageNet, Top-5 acc.) are marked as ``N/A''. 

For the  empirical results reported in Table~\ref{acc_ece}, we discuss generalization / test performance in   Section~\ref{generalization_expr} and calibration results in Section~\ref{calibration_expr}. 
Results for 121 tabular datasets are reported in Section~\ref{sec_121}.
We report the {\em average} accuracy/error rate (for test performance) and {\em average} ECE (for model calibration) of \textit{5 runs} with different random initializations for all experiments. 
We report the standard derivation of this collection of runs in Section~\ref{sec_std}.

\subsection{Empirical results on test performance}
\label{generalization_expr}

\begin{figure*}[ht]
    \centering
\centerline{\includegraphics[width=2.4\columnwidth]{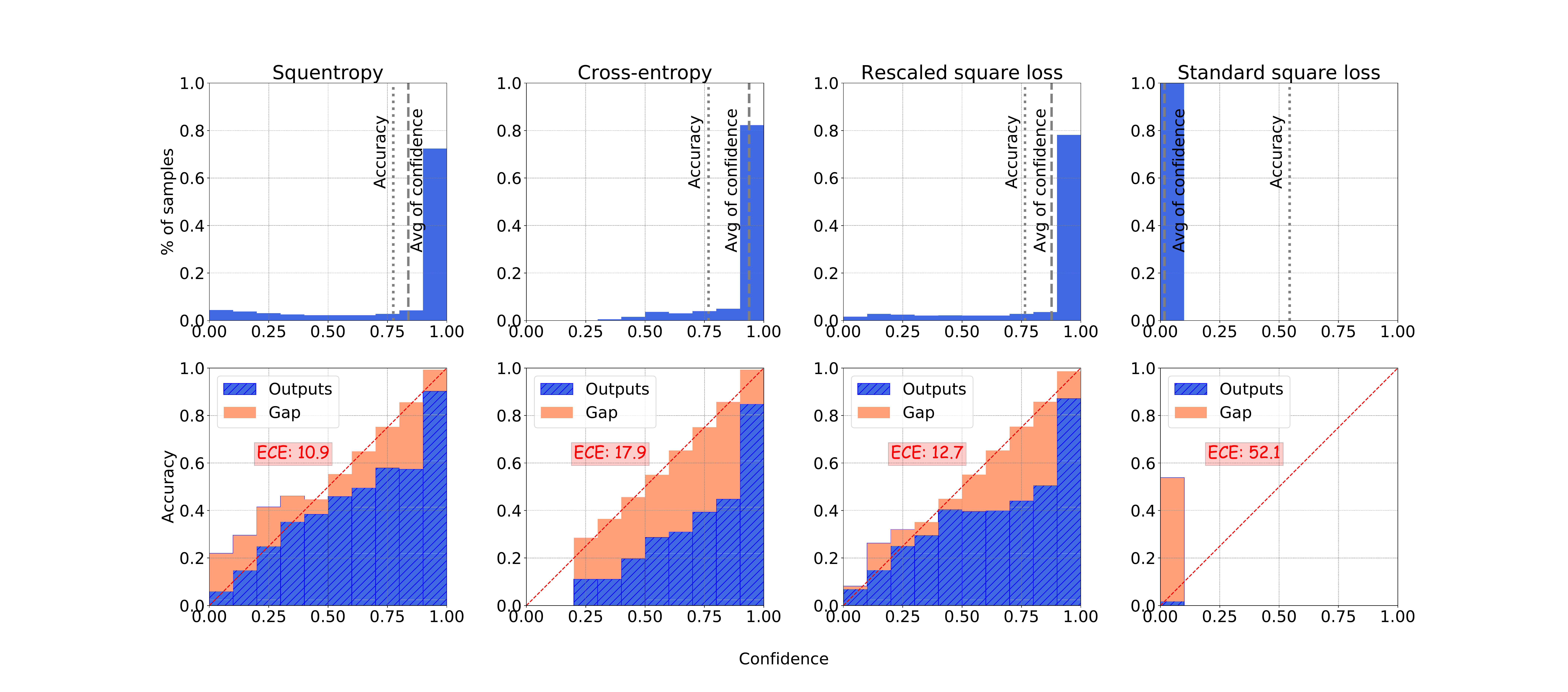}}
\vspace{-5mm}
    \caption{\textbf{Confidence histograms (top) and reliability diagrams (bottom) for a Wide Resnet on CIFAR-100.} See Table \ref{acc_ece} for its test accuracy. The Confidence histogram gives the portion of samples in each confidence interval, and the reliability diagrams show the accuracy as a function of the confidence. The ECE numbers are percentages as in Table \ref{acc_ece}. \textit{Left: }Squentropy, \textit{Middle left: } cross entropy, \textit{Middle right: } Rescaled square loss, \textit{Right: } Standard square loss. We see that models trained with squentropy are better calibrated, while cross entropy suffers from overconfidence and the standard square loss is highly underconfident.}
    \label{fig:diagram}
    \vspace{-3mm}
\end{figure*}

Our results show that squentropy has better test performance than  cross entropy and square loss in the majority of our experiments. 
The perf(\%) numbers in Table \ref{acc_ece} show the test accuracy of benchmarks of the NLP and vision tasks, and error rate for the speech tasks. 
Squentropy behaves the best in {\em 24 out of 34} tasks. 
We also report the numbers for {\em subsets} of enwik8 and CIFAR-100. 
Compared with full datasets of these collections, squentropy seems to gain more when the datasets are small.

\paragraph{Applicability and significance.}
Table \ref{acc_ece} shows  improvements for squentropy across a wide range distributions from the NLP, speech, and vision domains.
On the other hand, the improvement on one single task often is not significant, and for some datasets, squentropy's performance is worse.
One reason may be our choice to use the optimal hyperparameter values for cross entropy in squentropy. Further tuning of these hyperparameters may yield significant improvements.

\subsection{Empirical results on calibration}
\label{calibration_expr}

In this section we show model calibration results, measured with ECE of the models given in Table \ref{acc_ece}. The ECE numbers for NLP, speech, and vision tasks are also shown in Table \ref{acc_ece}. 
\paragraph{Squentropy consistently improves calibration.} As can be seen in and Table \ref{acc_ece}, in {26 out of 32} tasks, the calibration error (ECE) of models trained with squentropy is smaller than for cross entropy and  square loss, even in those  cases in which squentropy had slightly worse test performance, such as WSJ, STL10, and SVHN. 

Besides using ECE to measure model calibration, we also provide a popular form of visual representation of model calibration: reliability diagrams \cite{degroot1983comparison, niculescu2005predicting}, which show accuracy as a function of confidence as a bar chart. 
If the model is perfectly calibrated, i.e. $\mathbb{P}(\hat{y}_i=y_i|p_i)=p_i$, the diagram should show all bars aligning with the identity function. 
If most of the bars lie below the identity function, the model is overconfident as the confidence is mostly larger than corresponding accuracy. 
When most bars lie above the identity function that means the model is underconfident as confidence is smaller than accuracy. For a given bin $k$, the difference between $\text{acc}(B_k)$ and  $\text{conf}(B_k)$ represents the calibration \textit{gap} (orange bars in reliability diagrams – e.g. the bottom row of Figure \ref{fig:diagram}).

In Figure \ref{fig:diagram} we plot the confidence histogram (top) and the reliability diagrams (bottom) of Wide Resnets on CIFAR-100, trained with four different loss functions: squentropy, cross entropy, rescaled square loss (with $t=1, M=10$), and standard square loss ($t=1, M=1$). 
The confidence histogram gives the percentage of samples in each confidence interval, while the reliability diagrams show the test accuracy as a function of confidence.

In the reliability diagrams of Figure \ref{fig:diagram} bottom, the orange bars, which represent the confidence {\em gap}, start from the top of the blue (accuracy) bar. 
We show $\text{conf}(B_k) - \text{acc}(B_k)$ for all intervals in all reliability diagram plots. Note that for intervals where confidence is smaller than accuracy, the orange bars go down from the top of the blue bars, such as the one in the right bottom of Figure \ref{fig:diagram}. 
More reliability diagrams for other tasks are given in Appendix \ref{app_diag}.

\paragraph{Squentropy vs. cross entropy.} If we compare the diagrams of squentropy and cross entropy, the bars for squentropy are closer to the identity function;  cross entropy apparently yields more overconfident models.
The gap for squentropy is also smaller than cross entropy in most confidence intervals.
\begin{figure}[ht]
  \centering
\centerline{\includegraphics[width=1\columnwidth]{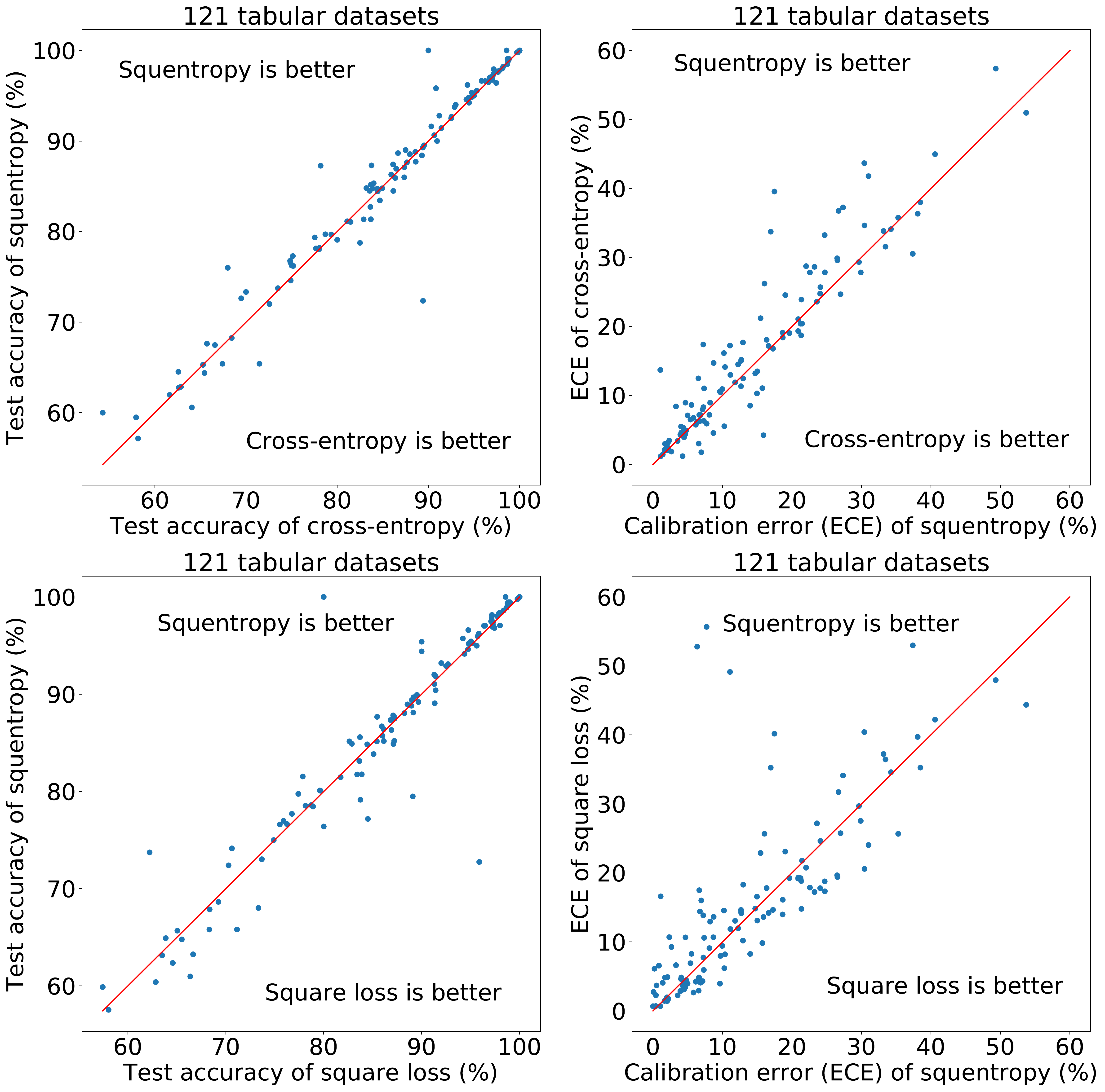}}
\vspace{1mm}
\resizebox{.45\textwidth}{!}{%
\begin{tabular}{cccc}
\hline
Loss functions & Squentropy      & Cross entropy & Square loss \\ \hline
Avg accuracy   & \textbf{85.6\%} & 85.2\%        & 85.5\%      \\
Avg ECE        & \textbf{11.6\%} & 13.0\%        & 15.7\%      \\ \hline
\end{tabular}}
\caption{\textbf{Test accuracy and model calibration of 121 tabular datasets from \cite{fernandez2014we}} trained with a 3 layer (64-128-64) fully connected network. The results for each dataset are averaged over 5 runs with different random initializations. \textit{Left:} Test accuracy (larger is better). \textit{Right:} Calibration error ECE (smaller is better). The top figures plot the results of squentropy and cross entropy, while the bottom figures plot the results of squentropy and the (rescaled) square loss. Test accuracy/ECE for each dataset are tabulated in Appendix \ref{app_121}.}

\label{121_results}
\vspace{-5mm}
\end{figure}
\paragraph{Standard square loss leads to underconfidence.} We also plot the reliability diagrams for training with the standard square loss on the right ones of Figure \ref{fig:diagram}. We see that it is highly underconfident as the confidence is smaller than $0.1$ (exact number is $0.017$) for all samples. Note that the square loss is directly applied to the logits $f_j(x_i)$, and the logits are driven to the one-hot vector $\bm{e}_{y_i}$, then the {\em probabilities} $p_{i,j}$ formed from these logits are not going to be close to the one-hot vector. The ``max'' probability (confidence) will instead be close to $\frac{e}{e+(C-1)}$, which is small when $C$ is large. 

\paragraph{Rescaling helps with calibration.} The second-from-right bottom diagram in Figure~\ref{fig:diagram} shows the results of training with the rescaled square loss ($t=1, M=10$) on CIFAR-100. This minimization problem drives the logits of true class closer to $M$, making the max probability approach $\frac{e^M}{e^M+(C-1)}$ - a much larger value than for standard square loss, leading to  better calibration.  However, squentropy can avoid extra rescaling hyperparameters while achieving even smaller values of ECE.

\subsection{Additional results on 121 Tabular datasets}
\label{sec_121}
Additional results for  121 small, low dimensional, and class-imbalanced tabular dataset, obtained with  3-layer fully-connected networks, are shown in  Figure~\ref{121_results}. 
For all these cases, we use SGD optimizer with weight decay parameter $5*10^{-4}$ and run $400$ epochs with learning rate $0.01$. The ``square loss'' function used here is in fact rescaled version with parameters $t=1$ and $M=5$. 

Figure~\ref{121_results} shows that for most datasets, squentropy has slightly  better test accuracy and significantly smaller ECE than cross entropy or square loss. 
Squentropy has the best test accuracy in 71 out of 121 tasks and best calibration in 60 tasks. 
If only compare with cross entropy, squentropy is better in 94 tasks on accuracy, and is better on calibration in 83 tasks. 
Test accuracy and ECE for each dataset in this collection are tabulated in Appendix~\ref{app_121}.

\input{variance}

\begin{figure*}[ht]
  \centering
  \begin{minipage}[b]{1.5\columnwidth}
\includegraphics[width=\columnwidth]{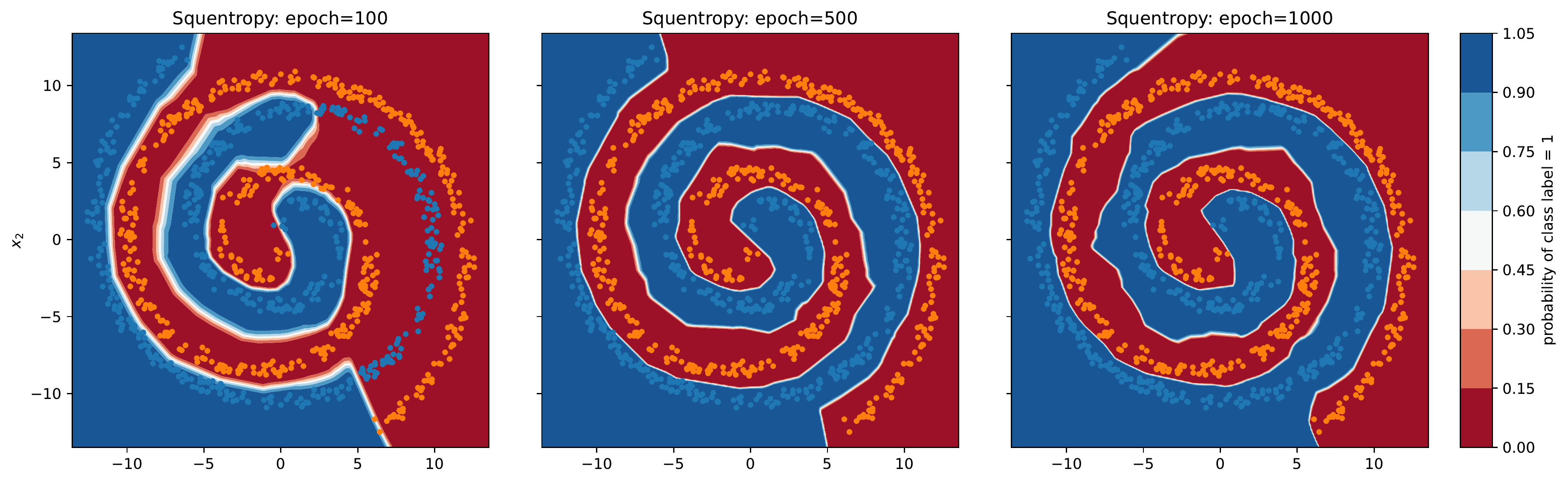}
  \end{minipage}
  \hfill
  \begin{minipage}[b]{1.5\columnwidth}
\includegraphics[width=\columnwidth]{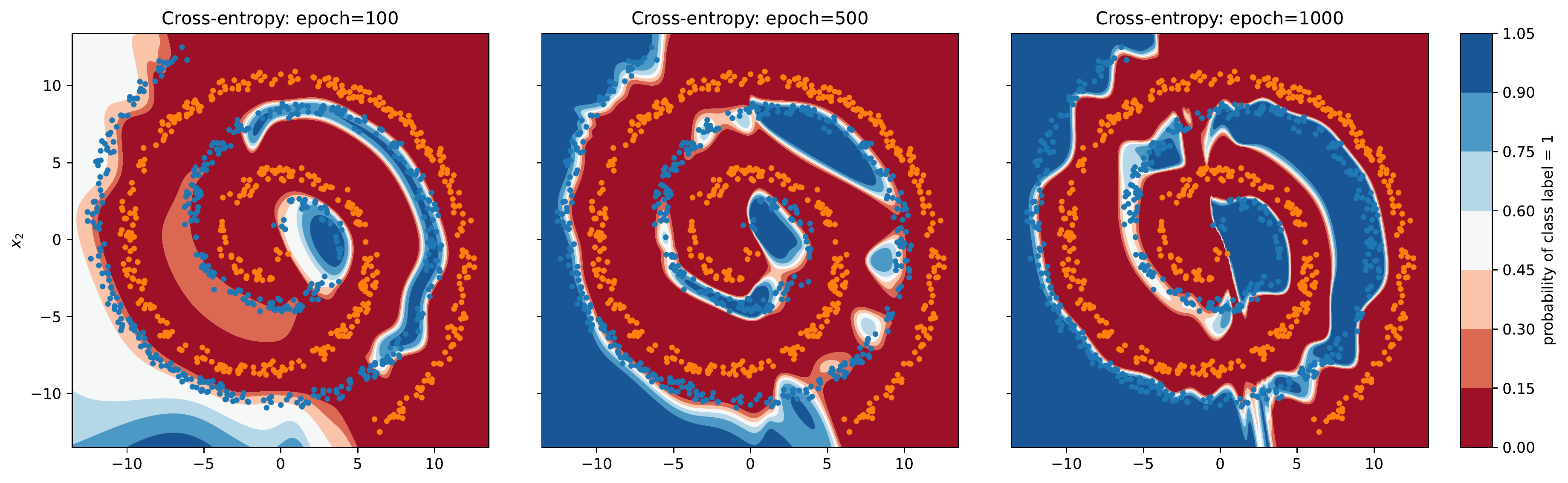}
  \end{minipage}
  \hfill
  \begin{minipage}[b]{1.5\columnwidth}
\includegraphics[width=\columnwidth]{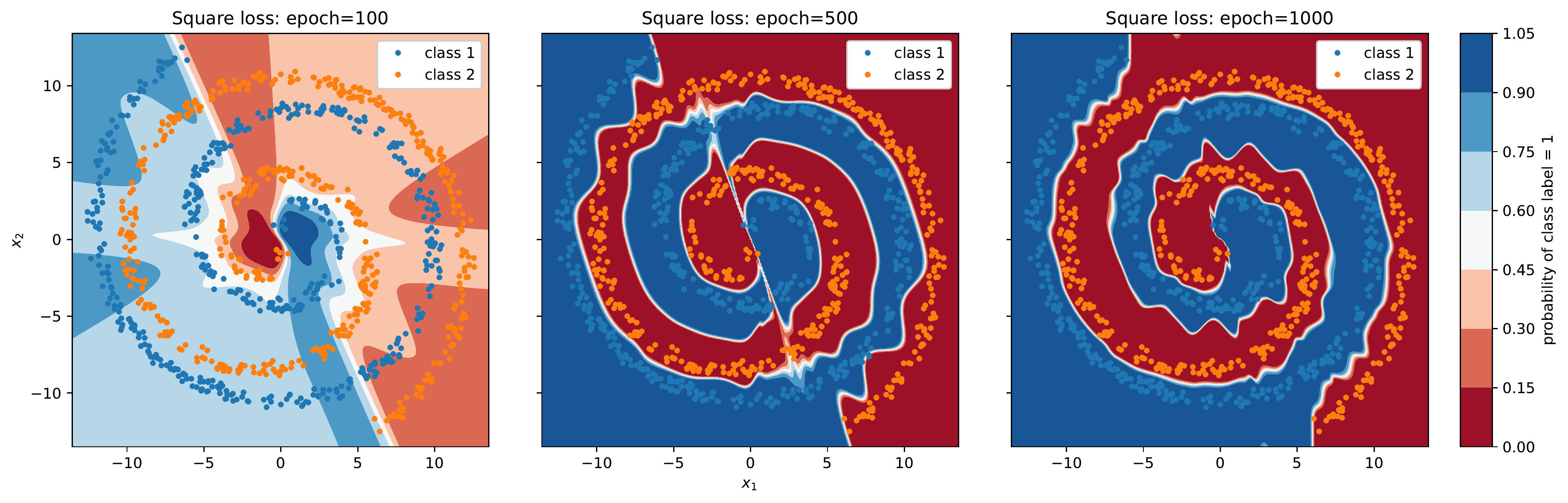}
  \end{minipage} 
  \vspace{-3mm}
  \caption{\textbf{Decision boundary along different epochs for test samples.} We fix all random seeds to be the same for all cases and hence the test set is exactly the same. (Thus, we display legends only in the bottom-row figures). Color coding indicates the calculated probability of class label to be $1$, according to the scale on th eright. The white line between red and blue areas indicates the decision boundary. We train a 3-layer fully connected network with 12 units in each layer, for a 2-class spiral data set in $\mathbb{R}^2$. There are 1000 samples for training and 500 samples for test, and we train for 1000 epochs, yielding a training accuracy of $100\%$ for all loss functions. Test accuracies are squentropy: $99.9\%$, cross entropy: $99.7\%$, square loss: $99.8\%$. 
  \textit{Top:} squentropy. \textit{Middle:} cross entropy. \textit{Bottom: }square loss. Columns show results after 100, 500, and 1000 epochs, respectively.}
  \label{fig_bd}
\end{figure*}
\subsection{Robustness to initialization}
\label{sec_std}
To evaluate the stability of the model trained with the loss functions considered in this paper, we report the standard deviation of the accuracy/error rate with respect to the randomness in initialization of weights for NLP, speech, and vision tasks. 
Standard deviation is over 5 runs with different random initializations; see Table~\ref{variance} for results.
The standard derivation of squentropy is smaller in the majority of the tasks considered, so results are comparatively insensitive to model initialization.

\section{Observations}
As mentioned previously, we conjecture that the square term of squentropy acts as an implicit regularizer and in this section we provide some observations in support of this conjecture. We discuss the decision boundary learnt by a fully-connected network on a 2-class spiral data problem (the ``Swiss roll") in Section~\ref{db_exp}, and remark on the weight norm of the last linear layer of several networks in Section~\ref{sec_norm}.

\subsection{Predicted probabilities and decision boundary}
\label{db_exp}
Using a simple synthetic setting, we observe that the decision boundary learned with squentropy appears to be  smoother than that for cross entropy and  the square loss.
We illustrate this point with a 2-class classification task with spiral data and a 3-layer fully-connected network with parameter $\theta$. This setup enables visual observations.
Given a sample $\bm{x}_i\in \mathbb{R}^2$ and labels $y_i \in \{1,2\}$, and the one hot encoding $\bm{y}_i=[0, 1]$ or $\bm{y}_i=[1, 0]$, we solve for weights $\theta$ to define functions $f_1(\bm{x}_i)$ and $f_2(\bm{x}_i)$ corresponding to the two classes.
For any $\bm{x}_i$, we then predict a probability of $\bm{x}_i$ being classified as class $1$ as follows: $p(\bm{x}_i) := e^{f_1(\bm{x}_i)} / (e^{f_1(\bm{x}_i)} + e^{f_2(\bm{x}_i)})$. 
Samples are assigned to class $1$ if $f_{i,1}>f_{i,2}$ and to class 2 otherwise. 
The decision boundary is the set of points for which $\{\bm{x} \, | \, f_{1}(\bm{x})=f_{2}(\bm{x}) \}$ or $\{\bm{x}|p(\bm{x}_i) = 1/2\}$.

We see from Figure~\ref{fig_bd} that the decision boundary obtained with squentropy is smoother than those learnt with both  cross entropy and square loss. This appears to be true throughout the training process, on this simple example.
Meanwhile, the margin (distance from training points to the decision boundary) is also larger for squentropy in many regions. Together, the large margin and smooth decision boundary imply immunity to perturbations and could be one of the reasons for the improved  generalization resulting from the use of squentropy~\cite{elsayed2018large}.

\begin{figure}[ht]
\centerline{\includegraphics[width=0.85\columnwidth]{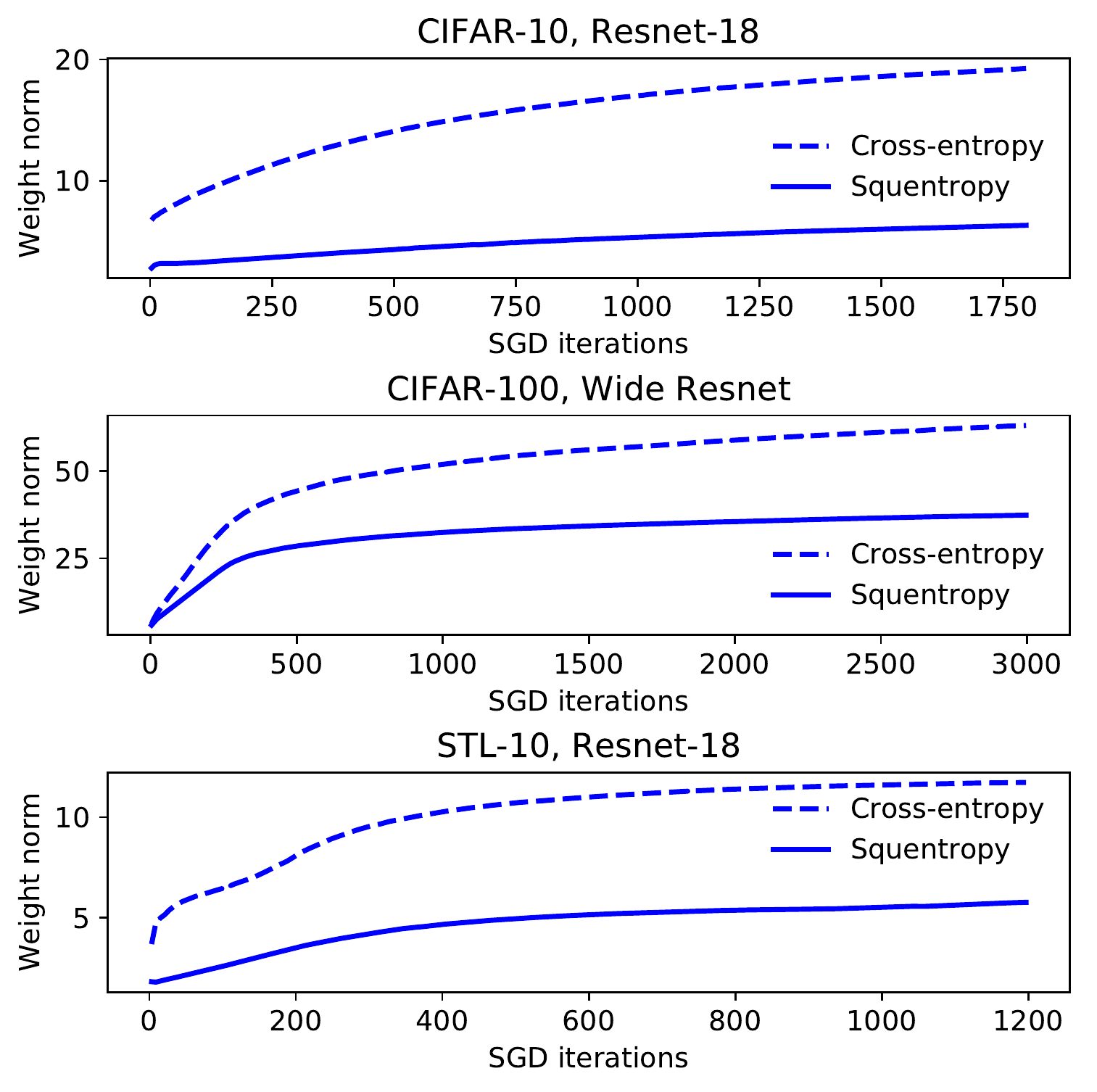}} 
\vspace{-5mm}
\caption{\textbf{Weight norm along training.} We train a Resnet-18 on CIFAR-10 (calibration error, ECE: Squentropy: $8.9\%$, cross entropy: $10.0\%$) and STL-10 (ECE: Squentropy: $21.2\%$, cross entropy: $26.1\%$), a wide Resnet on CIFAR-100 (ECE: Squentropy: $10.9\%$, cross entropy: $17.9\%$), and show the norm of the last linear layer's weights. These are the same experiments as given in Table \ref{acc_ece}.}
\label{w_norm}
\end{figure}
\vspace{-3mm}
\subsection{Weight norm}
\label{sec_norm}
Neural classifiers trained with cross-entropy loss suffer from overconfidence, causing miscalibration of the model \cite{guo2017calibration}. 
Our calibration results in Figure~\ref{fig:diagram} and Section~\ref{sec_more_reliability} show evidence of this phenomenon.
As can be seen in the confidence histogram of cross entropy --- the $(1,2)$ figure in  Figure~\ref{fig:diagram} --- the average confidence $p_{y_i}(\bm{x}_i)$ for the predicted label in cross entropy is close to $1$. 
This fact suggests that the logits $f_{y_i}(\bm{x}_i)$ of true class are close to $\infty$, while the logits of the incorrect classes approach $-\infty$. 
Such limits are possible only  when the weights of last linear layer have large norm. 
To quote \cite{mukhoti2020calibrating}, ``{\it cross-entropy loss thus inherently induces this tendency of weight magnification in neural network optimisation}.''

\citet{guo2017calibration} comment that weight decay, which corresponds to adding a penalty term to the loss consisting of the sum of squares of the weights, can produce appreciably better calibration while having a minimal effect on test error; see the rightmost diagram in Figure 2 of \cite{guo2017calibration}.
In \cite{mukhoti2020calibrating, liu2022devil}, the authors point out how focal loss proposed in~\cite{lin2017focal} improves calibration by encouraging the predicted distribution to have higher entropy, thus implicitly regularizing the weights. 
Figure~C.1 of \cite{mukhoti2020calibrating} compares weight norm and final logit values between cross entropy and the focal loss, showing that the latter are significantly smaller.
We perform a similar experiment, showing in Figure~\ref{w_norm} the weight norm of the final-layer weights for three examples from Table~\ref{acc_ece} as a function of training steps. We observe that the weight norm for the model trained with squentropy is much smaller than the norms for the same set of weights in the model trained with cross entropy, along the whole training process.

\section{Summary, thoughts, future investigations}

As with the selection of an optimization procedure, the choice of the loss function is  an ineluctable aspect of training  all modern neural networks. Yet the machine learning community has paid little attention to understanding the properties of loss functions. There is little justification, theoretical or empirical,  for the predominance of cross-entropy loss in practice. 
Recent work~ \citep{hui2020evaluation} showed that the square loss, which is universally used in regression, can perform at least as well as cross entropy in classification. 
Other  works have made similar observations: \cite{rifkin2002everything, sangari2015convergence, que2016back, demirkaya2020exploring}. While several alternative loss functions, such as the focal loss~\cite{lin2017focal},  have been considered in the literature with good results, none have been adopted widely. 
Even the hinge loss, the former leader in the popularity contest for classification losses, is barely used outside the context of Support Vector Machines.

In this work we demonstrate that a simple hybrid loss function can achieve better accuracy and better calibration than the standard  cross entropy on a significant  majority of a broad range of classification tasks. Our squentropy loss function has no tunable parameters. Moreover, most of our experiments were conducted in a true ``plug-and-play'' setting using the  same algorithmic parameters in the optimization process as for training with the  standard cross-entropy loss. 
Performance of squentropy can undoubtedly be further improved by tuning the optimization parameters. Furthermore, various calibration techniques can potentially be applied with squentropy in the same way they are used with cross entropy. 

Thus, from a practical point of view, squentropy currently appears to be the natural first choice to train neural models. 

By no means does it imply that we know of fundamental reasons or compelling intuition indicating that squentropy is the last word on the choice of loss functions for classification. One of the main goals of this work is to encourage both practitioners and theoreticians to investigate the properties of  loss functions, an important but largely overlooked aspect of modern Machine Learning. 

\section*{Acknowledgements}
We acknowledge support from the National Science Foundation (NSF) and the Simons Foundation for the Collaboration on the Theoretical Foundations of Deep Learning\footnote{\url{https://deepfoundations.ai/}} through awards DMS-2031883 and \#814639 as well as the  TILOS institute (NSF CCF-2112665). 
This work was supported also by an NSF TRIPODS grant to the Institute for Foundations of Data Science (NSF DMS-2023239), NSF grant CCF-222421, and AFOSR via subcontract UTA20-001224 from UT-Austin. 

LH thanks Chaoyue Liu and Parthe Pandit for reading the draft and give useful comments on the writing.
We thank Nvidia for the donation of GPUs and Google for providing access to the cloud TPUs. This work uses CPU/GPU nodes (allocated with TG-CIS220009) provided by San Diego Supercomputer center, with the Extreme Science and Engineering
Discovery Environment (XSEDE) \cite{towns2014xsede}, which is supported by NSF grant number ACI-1548562.


\bibliography{icml2023}
\bibliographystyle{icml2023}

%%%%%%%%%%%%%%%%%%%%%%%%%%%%%%%%%%%%%%%%%%%%%%%%%%%%%%%%%%%%%%%%%%%%%%%%%%%%%%%
%%%%%%%%%%%%%%%%%%%%%%%%%%%%%%%%%%%%%%%%%%%%%%%%%%%%%%%%%%%%%%%%%%%%%%%%%%%%%%%
% APPENDIX
%%%%%%%%%%%%%%%%%%%%%%%%%%%%%%%%%%%%%%%%%%%%%%%%%%%%%%%%%%%%%%%%%%%%%%%%%%%%%%%
%%%%%%%%%%%%%%%%%%%%%%%%%%%%%%%%%%%%%%%%%%%%%%%%%%%%%%%%%%%%%%%%%%%%%%%%%%%%%%%
\newpage
\appendix
\onecolumn
\section{Datasets}
\label{app_data}
Datasets used in our tests include the following.
\begin{itemize}
    \item CIFAR-100: \cite{krizhevsky2009learning} consists of 50, 000 32×32 pixel training images
and 10, 000 32 × 32 pixel test images in 100 different classes. It is a balanced dataset with
6, 00 images of each class.
    \item SVHN: \cite{netzer2011reading} is a real-world image dataset obtained from house numbers in Google Street View images and it incorporates over 600,000 digit images with labeles.  It is a good choice for developing machine learning and object recognition algorithms with minimal requirement on data preprocessing and formatting.
    \item STL-10: \cite{coates2011analysis} is an image recognition dataset mainly for developing unsupervised feature learning as it contains many images without labels. The resolution of this dataset is 96x96 and this makes it a challenging benchmark.
    
\end{itemize}

See Appendix A of \cite{hui2020evaluation} for details of other datasets.
\input{para}

\section{Hyperparameters}
\label{app_para}

Detailed hyperparameter settings for CIFAR-100, SVHN, and STL-10  are shown in Table \ref{vision_para}.
For the other tasks, we follow the exact same settings as provided in Appendix B of \cite{hui2020evaluation}.

\section{More reliability diagrams}
\label{sec_more_reliability}

We provide the reliability diagrams for more tasks. 
Note that the values given for ECE (Expected calibration error as defined in \eqref{eq:ece} and the smaller the better) in these plots are percentages as in Table~\ref{acc_ece}.

\label{app_diag}
\input{ece_plots}

\section{Results for 121 tabular datasets}
\label{app_121}
We list the test accuracy and calibration results (ECE) of each tabular dataset in Tables~\ref{121t1}, \ref{121t2} and \ref{121t3}. Note that the square loss of in those tables are all rescaled square loss defined in Equation (\ref{square_func}). with $t=1, M=5$.
\input{121_table_results}

%\input{ece_plots}

%%%%%%%%%%%%%%%%%%%%%%%%%%%%%%%%%%%%%%%%%%%%%%%%%%%%%%%%%%%%%%%%%%%%%%%%%%%%%%%
%%%%%%%%%%%%%%%%%%%%%%%%%%%%%%%%%%%%%%%%%%%%%%%%%%%%%%%%%%%%%%%%%%%%%%%%%%%%%%%

\end{document}

%% file: accuracy_ece.tex
\begin{table*}[ht]
\centering
\vspace{-3mm}
\caption{Test performance (\textbf{perf}(\%): accuracy for NLP\&Vision, error rate for speech data) and calibration: \textbf{ECE}(\%).}
\label{acc_ece}
%\begin{threeparttable}

\begin{tabular}{ccccccccc}
\hline
\hline
\multirow{2}{*}{Domain}   & \multirow{2}{*}{Model}                               & \multirow{2}{*}{Task}            & \multicolumn{2}{c}{Squentropy} & \multicolumn{2}{c}{Cross-entropy} &\multicolumn{2}{c}{Square loss}  \\ \cline{4-9}
&         &              & \multicolumn{1}{c|}{perf} & ECE & \multicolumn{1}{c|}{perf} &ECE &  \multicolumn{1}{c|}{perf} &ECE \\ \hline
\multirow{15}{*}{NLP} & \multirow{6}{*}{\begin{tabular}[c]{@{}c@{}}fine-tuned BERT \\\cite{devlin2018bert}\end{tabular}} & MRPC &  \multicolumn{1}{c|}{$\textbf{84.0}$}&$\textbf{7.9}$ &    \multicolumn{1}{c|}{$82.1$}& $13.1$   & \multicolumn{1}{c|}{$83.8$}&$14.0$\\
                              &    & SST-2   &   \multicolumn{1}{c|}{$\textbf{94.2}$}&$7.0$               & \multicolumn{1}{c|}{$93.9$}&$\textbf{6.7}$  & \multicolumn{1}{c|}{$94.0$}&$19.8$ \\
                             &     & QNLI     &  \multicolumn{1}{c|}{$\textbf{91.0}$}&$7.3$         & \multicolumn{1}{c|}{$90.6$}&$7.4$  & \multicolumn{1}{c|}{$90.6$ }&$\textbf{4.2}$\\
                           &       & QQP &    \multicolumn{1}{c|}{$\textbf{89.0}$} & $\textbf{2.2}$        & \multicolumn{1}{c|}{$88.9$}&$5.8$   &\multicolumn{1}{c|}{$88.9$} & $2.8$ \\
                           &       & text5 & \multicolumn{1}{c|}{$\textbf{85.2}$}&$\textbf{12.4}$ &\multicolumn{1}{c|}{$84.5$}&$14.9$     & \multicolumn{1}{c|}{$84.6$}&$46.7$\\
                           &       & text20   &  \multicolumn{1}{c|}{$\textbf{81.2}$}&$\textbf{10.5}$                 & \multicolumn{1}{c|}{$80.8$}&$16.2$ & \multicolumn{1}{c|}{$80.8$}&$69.2$ \\
                                  \cline{2-9}

& \multirow{3}{*}{\begin{tabular}[c]{@{}c@{}}Transformer-XL \\ \cite{dai2019transformer}\end{tabular}} & text8 &  \multicolumn{1}{c|}{$71.5$}&$\textbf{3.9}$  &    \multicolumn{1}{c|}{$\textbf{72.8}$}&$5.8$   & \multicolumn{1}{c|}{$73.2$}&$57.6$\\
                          &        & enwik8   &    \multicolumn{1}{c|}{$77.0$}& $\textbf{4.8}$           & \multicolumn{1}{c|}{$\textbf{77.5}$}&$9.3$  & \multicolumn{1}{c|}{$76.7$}&$64.5$ \\
                         &         & enwik8 (subset) & \multicolumn{1}{c|}{$\textbf{48.9}$}&$\textbf{10.7}$  & \multicolumn{1}{c|}{$48.6$}&$18.9$ & \multicolumn{1}{c|}{$47.3$}&$70.6$  \\
                                   \cline{2-9}
&\multirow{3}{*}{\begin{tabular}[c]{@{}c@{}}LSTM+Attention  \\ \cite{chen2016enhanced} \end{tabular}}              & MRPC&   \multicolumn{1}{c|}{$71.4$}&$\textbf{3.2}$ & \multicolumn{1}{c|}{$70.9$}&$7.1$   & \multicolumn{1}{c|}{$\textbf{71.7}$}&$3.5$\\
                          &        & QNLI &  \multicolumn{1}{c|}{$\textbf{79.3}$}&$\textbf{7.2}$    & \multicolumn{1}{c|}{$79.0$}&$7.6$     &\multicolumn{1}{c|}{$\textbf{79.3}$}&$13.0$ \\
                            &      & QQP  &  \multicolumn{1}{c|}{$\textbf{83.5}$}&$\textbf{2.4}$    & \multicolumn{1}{c|}{$83.1$}&$3.2$        &    \multicolumn{1}{c|}{ $\textbf{83.4}$}&$16.5$ \\ \cline{2-9}
&\multirow{3}{*}{\begin{tabular}[c]{@{}c@{}}LSTM+CNN \\ \cite{he2016pairwise}\end{tabular}}          & MRPC&    \multicolumn{1}{c|}{$70.5$}&$\textbf{5.2}$               &      \multicolumn{1}{c|}{ $69.4$} & $6.3$  &   \multicolumn{1}{c|}{$\textbf{73.2}$}&$16.3$   \\
                          &        & QNLI &  \multicolumn{1}{c|}{$\textbf{76.0}$}&$4.1$                   &    \multicolumn{1}{c|}{$\textbf{76.0}$}&$\textbf{2.3}$ & \multicolumn{1}{c|}{$\textbf{76.0}$}&$20.5$ \\&
                                  & QQP  &   \multicolumn{1}{c|}{$\textbf{84.5}$}&$\textbf{5.1}$                &   \multicolumn{1}{c|}{$84.4$}&$7.2$ &\multicolumn{1}{c|}{ $84.3$}&$24.6$ \\ \hline

\multirow{8}{*}{Speech} &\multirow{2}{*}{\begin{tabular}[c]{@{}c@{}} Attention+CTC \\ \cite{kim2017joint} \end{tabular}} & TIMIT (PER)  &  \multicolumn{1}{c|}{$\textbf{19.6}$}&$\textbf{0.7}$ & \multicolumn{1}{c|}{$20.0$}&$3.1$  &    \multicolumn{1}{c|}{$20.0$}&$2.8$       \\
& & TIMIT (CER)  & \multicolumn{1}{c|}{$\textbf{32.1}$}&$\textbf{1.6}$   & \multicolumn{1}{c|}{$33.4$}&$3.3$ &    \multicolumn{1}{c|}{$32.5$}&$4.3$       \\
&\multirow{2}{*}{\begin{tabular}[c]{@{}c@{}} VGG+BLSTMP  \\ \cite{moritz2019triggered}
 \end{tabular}}  & WSJ (WER) & \multicolumn{1}{c|}{$5.5$}&$\textbf{3.2}$   &  \multicolumn{1}{c|}{$5.3$}&$5.0$ & \multicolumn{1}{c|}{$\textbf{5.1}$}&$5.3$ \\
 & & WSJ (CER) &  \multicolumn{1}{c|}{$2.9$}&$\textbf{3.2}$   & \multicolumn{1}{c|}{$2.5$}&$5.0$ & \multicolumn{1}{c|}{$\textbf{2.4}$}&$5.3$ \\
& \multirow{2}{*}{\begin{tabular}[c]{@{}c@{}} VGG+BLSTM \\ 
\cite{moritz2019triggered} \end{tabular}}& Librispeech (WER)   & \multicolumn{1}{c|}{$\textbf{7.6}$}&$7.1$    & \multicolumn{1}{c|}{$8.2$}&$\textbf{2.7}$ & \multicolumn{1}{c|}{ $8.0$}&$7.9$  \\ 
& & Librispeech (CER)  & \multicolumn{1}{c|}{$\textbf{9.7}$}&$ 7.1 $    &  \multicolumn{1}{c|}{$10.6$}&$ \textbf{2.7} $  &  \multicolumn{1}{c|}{$\textbf{9.7}$}&$ 7.9 $ \\
% & & Librispeech (SER)  & \multicolumn{1}{c|}{$\textbf{69.2}$}&$-$    & \multicolumn{1}{c|}{ $71.5$}&$-$  &  \multicolumn{1}{c|}{$69.4$}&$-$ \\
& \multirow{2}{*}{\begin{tabular}[c]{@{}c@{}} Transformer \\ \cite{watanabe2018espnet} \end{tabular}}    & WSJ (WER)         &  \multicolumn{1}{c|}{$\textbf{3.9}$}&$\textbf{2.1}$ &  \multicolumn{1}{c|}{$4.2$}&$4.3$  &  \multicolumn{1}{c|}{$4.0$}&$4.4$     \\
& & Librispeech (WER)  &\multicolumn{1}{c|}{$\textbf{9.1}$}&$\textbf{4.2}$     & \multicolumn{1}{c|}{$9.2$}&$4.9$ & \multicolumn{1}{c|}{$9.4$}&$5.1$  \\  \hline
\multirow{11}{*}{Vision} & TCNN \cite{bai2018empirical}   & MNIST      &    \multicolumn{1}{c|}{$\textbf{97.8}$}&$\textbf{1.4}$   & \multicolumn{1}{c|}{$97.7$}&$1.6$    &       \multicolumn{1}{c|}{$97.7$}&$75.0$   \\ 
&\multirow{2}{*}{\begin{tabular}[c]{@{}c@{}} Resnet-18 \\ \cite{he2016deep} \end{tabular}} & CIFAR-10  & \multicolumn{1}{c|}{$\textbf{85.5}$}&$\textbf{8.9}$  & \multicolumn{1}{c|}{$84.7$}&$10.0$  & \multicolumn{1}{c|}{$84.6$}&$13.4$ \\ 
& & STL-10 &  \multicolumn{1}{c|}{$93.0$}&$\textbf{21.2}$ & \multicolumn{1}{c|}{$\textbf{93.7}$}&$26.1$  & \multicolumn{1}{c|}{$92.5$}&$40.3$ \\
&\multirow{2}{*}{\begin{tabular}[c]{@{}c@{}} W-Resnet \\ \cite{zagoruyko2016wide}\end{tabular}} & CIFAR-100 & \multicolumn{1}{c|}{$\textbf{77.5}$} & $\textbf{10.9}$ & \multicolumn{1}{c|}{$76.7$}  &$17.9$ & \multicolumn{1}{c|}{$76.5$ }&$12.7$\\
& & CIFAR-100 (subset) & \multicolumn{1}{c|}{$\textbf{43.5}$}&$\textbf{18.8}$ & \multicolumn{1}{c|}{$41.5$}&$40.3$ & \multicolumn{1}{c|}{$41.0$}&$23.8$ \\
&Visual transformer  & CIFAR-10  & \multicolumn{1}{c|}{$\textbf{99.3}$}&$\textbf{1.9}$   & \multicolumn{1}{c|}{$99.2$}&$3.8$  &   \multicolumn{1}{c|}{$\textbf{99.3}$}&$7.2$\\ 
&VGG & SVHN  & \multicolumn{1}{c|}{$67.7$}&$\textbf{4.8}$ & \multicolumn{1}{c|}{$\textbf{68.9}$}&$5.7$  & \multicolumn{1}{c|}{$67.0$}&$65.4$ \\

\cline{2-9}
&\multirow{2}{*}{\begin{tabular}[c]{@{}c@{}} Resnet-50 \\ \cite{he2016deep}\end{tabular}}    & ImageNet (acc.)       & \multicolumn{1}{c|}{ $\textbf{76.3}$}& $\textbf{6.3}$  & \multicolumn{1}{c|}{$76.1$}&$6.7$  &     \multicolumn{1}{c|}{ $ 76.2$}& 8.2 \\
& & ImageNet (Top-5 acc.) &\multicolumn{1}{c|}{$ \textbf{93.2}$} &N/A & \multicolumn{1}{c|}{$93.0$}&N/A &     \multicolumn{1}{c|}{ $93.0$}&N/A \\
& \multirow{2}{*}{\begin{tabular}[c]{@{}c@{}}EfficientNet \\ \cite{tan2019efficientnet}\end{tabular}} & ImageNet (acc.)       &  \multicolumn{1}{c|}{$76.4$}&  $6.8$  & \multicolumn{1}{c|}{$\textbf{77.0}$}&   $\textbf{5.6}$  & \multicolumn{1}{c|}{$74.6$}&   $7.9$   \\
& & ImageNet (Top-5 acc.) & \multicolumn{1}{c|}{$93.0$}&N/A   & \multicolumn{1}{c|}{$\textbf{93.3}$}&N/A&     \multicolumn{1}{c|}{$92.7$}&N/A\\ \hline
\end{tabular}
% \begin{tablenotes}[para,flushleft]
% \label{TER}
% \item [$*$] For WSJ, WER shares the same acoustic model with CER, the calibration numbers are the same for CER. Similarly, WER, CER and SER of Librispeech shares the same acoustic model, hence only one calibration results. 
% \item[$^\diamondsuit$]For Top-5 accuracy,  the ECE numbers are not available.
% \end{tablenotes}
% \end{threeparttable}
\vspace{-2mm}
\end{table*}

%% file: variance.tex
\begin{table}[!htbp]
\centering
\vspace{-4mm}
\caption{Standard deviation of test accuracy/error. Smaller number is bolded. CE is short for cross-entropy.} 
\label{variance}
\resizebox{.88\linewidth}{!}{%
\begin{tabular}{ccccc}
\hline
\hline
Model                           & Dataset       &Squentropy    & CE  & Square loss             \\ \hline
\multirow{6}{*}{fine-tuned BERT}           &   MRPC          &   \textbf{0.285}   & 0.766              &     0.484           \\
                                &    SST-2 &   \textbf{0.144}           &  0.173               &      0.279          \\
                                &  QNLI    &   \textbf{0.189}          &  0.205
             &      0.241         \\
                                &  QQP     &  0.050          & 0.063               &  \textbf{0.045}              \\
                                &  text5   &    \textbf{0.132}          &   0.167            &   0.147             \\
                                &  text20  &  0.131             &       \textbf{0.08}        & 0.172               \\ \hline
\multirow{2}{*}{\begin{tabular}[c]{@{}c@{}}Transformer-XL\end{tabular}} & text8      &   \textbf{0.149}     & 0.204    &     0.174      \\
                                & enwik8   &    0.156       & \textbf{0.102} &    0.228      \\
                                \hline

\multirow{3}{*}{\begin{tabular}[c]{@{}c@{}}LSTM\\+Attention\end{tabular}} & MRPC        &  \textbf{0.315}    & 0.786 & 0.484          \\
                                & QNLI     &   \textbf{0.198}      & 0.371 & 0.210          \\
                                & QQP      &   0.408      & \textbf{0.352}           & 0.566 \\ \hline
\multirow{3}{*}{\begin{tabular}[c]{@{}c@{}}LSTM\\+CNN\end{tabular}}       & MRPC        &   \textbf{0.289}   & 0.383 & 0.322          \\
                                & QNLI     &   \textbf{0.154}     & 0.286  & 0.173         \\
                                & QQP      &    0.279     & \textbf{0.161}           & 0.458 \\ \hline
\multirow{2}{*}{\begin{tabular}[c]{@{}c@{}}Attention\\+CTC\end{tabular}}  & TIMIT (PER)   & 0.332   & \textbf{0.249}          & 0.508 \\
                                & TIMIT (CER)  &  \textbf{0.232}   & 0.873  & 0.361          \\ \hline
\multirow{2}{*}{\begin{tabular}[c]{@{}c@{}}VGG+\\BLSTMP\end{tabular}}     & WSJ (WER)   &  \textbf{0.147}    & 0.249  & 0.184         \\
                                & WSJ (CER)      &  0.082 & 0.118 & \textbf{0.077}          \\ \hline
\multirow{2}{*}{\begin{tabular}[c]{@{}c@{}}VGG+\\BLSTM\end{tabular}}      & Libri (WER)& \textbf{0.117}  & 0.257  & 0.126          \\
                                & Libri (CER) & \textbf{0.125}&0.316  & 0.148           \\
\multirow{2}{*}{\begin{tabular}[c]{@{}c@{}} Transformer\end{tabular}}      & WSJ (WER) &  \textbf{0.186}    & 0.276    &    0.206   \\
                                & Libri (WER) &  0.168 &  0.232  &   \textbf{0.102}           \\                                
                                \hline
TCNN                            & MNIST          &  \textbf{0.151} & 0.173  & 0.161          \\ \hline
\multirow{2}{*}{\begin{tabular}[c]{@{}c@{}} Resnet-18\end{tabular}}     & CIFAR-10     & \textbf{0.147}  &   0.452 &      0.174     \\ 
& STL-10 &   0.413   &   0.376    &  \textbf{0.230} \\ \hline
W-ResNet      & CIFAR-100     &  \textbf{0.164}   &   0.433   &    0.181    \\ \hline
Visual Transformer                     & CIFAR-10    &  0.070    &  0.075     &  \textbf{0.063}         \\ \hline
VGG  & SVHN & 0.283   &   \textbf{0.246}      & 0.307 \\ \hline
\multirow{2}{*}{Resnet-50}      & I-Net (Top-1)  &    \textbf{0.029}    & 0.045 & 0.032          \\
                                & I-Net (Top-5)         & 0.098 &\textbf{0.045}          & 0.126 \\ \hline
\multirow{2}{*}{EfficientNet}   & I-Net (Top-1)  &   \textbf{0.099}     & 0.122          & 0.138 \\
                                & I-Net (Top-5)    & 0.092    &\textbf{0.089}  & \textbf{0.089} \\ \hline
\end{tabular}
}%
\vspace{-4mm}
\end{table}

%% file: para.tex
\begin{table}[!ht]
\centering
\vspace{-2mm}
\caption{Hyper-parameters for CIFAR-100, SVHN, and STL-10.}
\label{vision_para}
\begin{threeparttable}
\begin{tabular}{|c|c|c|c|c|c|}
\hline
\multirow{2}{*}{Model} & \multirow{2}{*}{Task} & \multirow{2}{*}{Hyper-parameters} & \multicolumn{3}{c|}{Epochs training w/} \\ \cline{4-6} 
                       &                       &                           &squentropy        & square loss           & CE              \\ \hline

Wide-ResNet            & CIFAR-100              &   \begin{tabular}[c]{@{}c@{}}   lr=0.1, layer=28 \\ wide-factor=20, batch size: 128    \end{tabular}                         & 200            &200       & 200             \\ \hline
VGG             & SVHN              &   \begin{tabular}[c]{@{}c@{}} lr=0.1 for cross-entropy\\lr=0.0.02 for squentropy and square loss\end{tabular}                                & 200                & 200 &200         \\ \hline
Resnet-18           & STL-10              & \begin{tabular}[c]{@{}c@{}} lr=0.1 for cross-entropy\\ for squentropy and square loss lr=0.02                 \end{tabular}               & 200 &200              & 200           \\ \hline
\end{tabular}
\end{threeparttable}
\vspace{-4mm}
\end{table}

%% file: ece_plots.tex
\begin{figure*}[!htbp]
  % \centering
\centerline{\includegraphics[width=1\columnwidth]{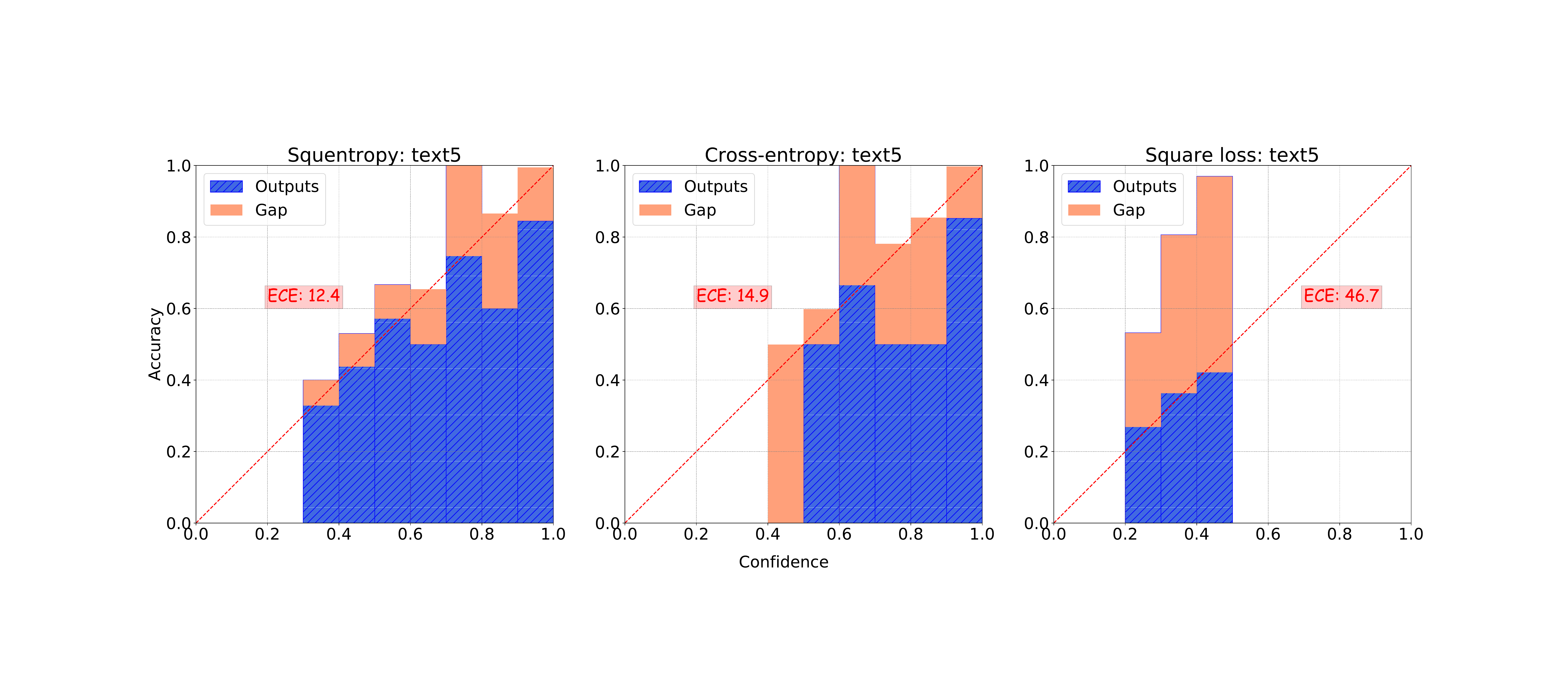}} 
\vspace{-15mm}
\caption{Reliability diagrams for a pretrained BERT
on text5 data. \textit{Left:} squentropy, \textit{middle:} cross-entropy, \textit{right:} square loss.}
\end{figure*}
\begin{figure*}[!htbp]
  % \centering
\centerline{\includegraphics[width=1\columnwidth]{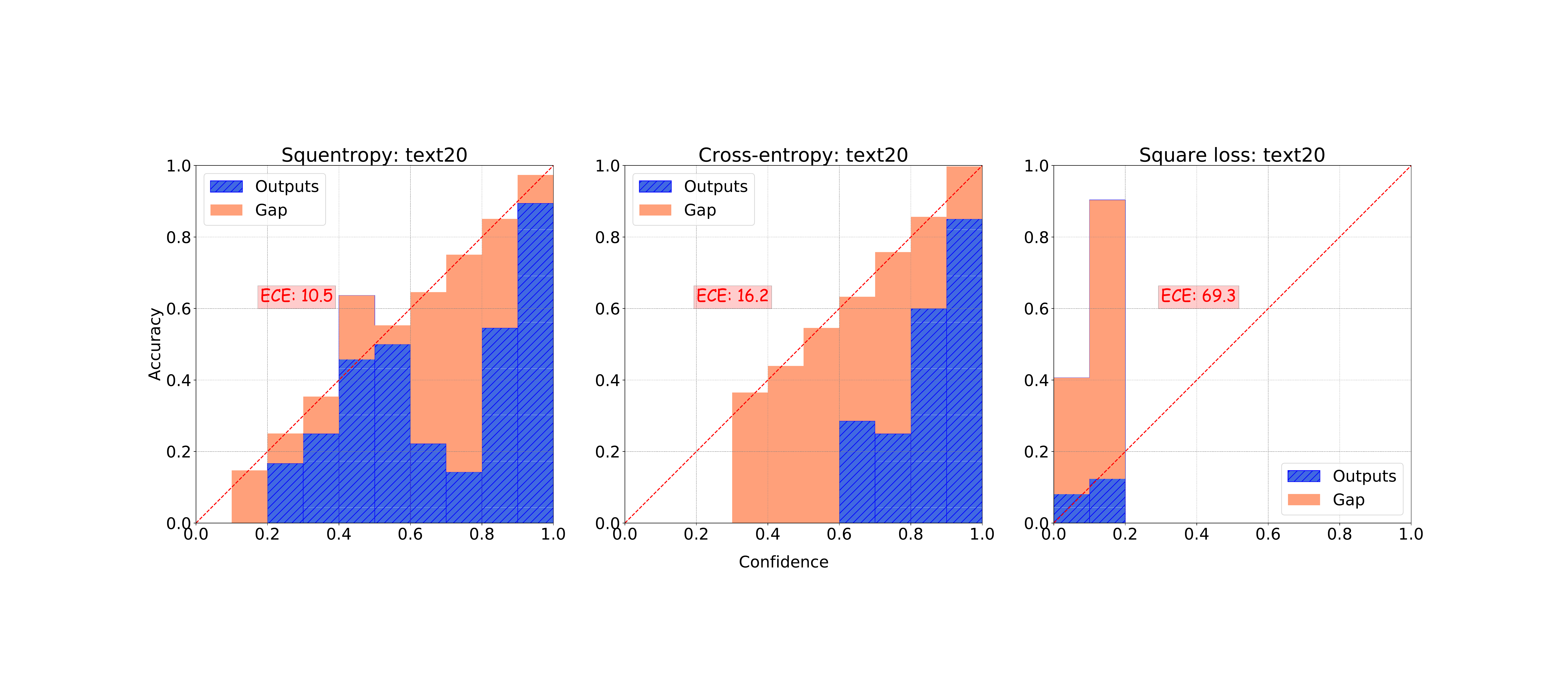}} 
\vspace{-15mm}
\caption{Reliability diagrams for a pretrained BERT
on text20 data. \textit{Left:} squentropy, \textit{middle:} cross-entropy, \textit{right:} square loss.}
\end{figure*}
\begin{figure*}[!htbp]
  % \centering
\centerline{\includegraphics[width=1\columnwidth]{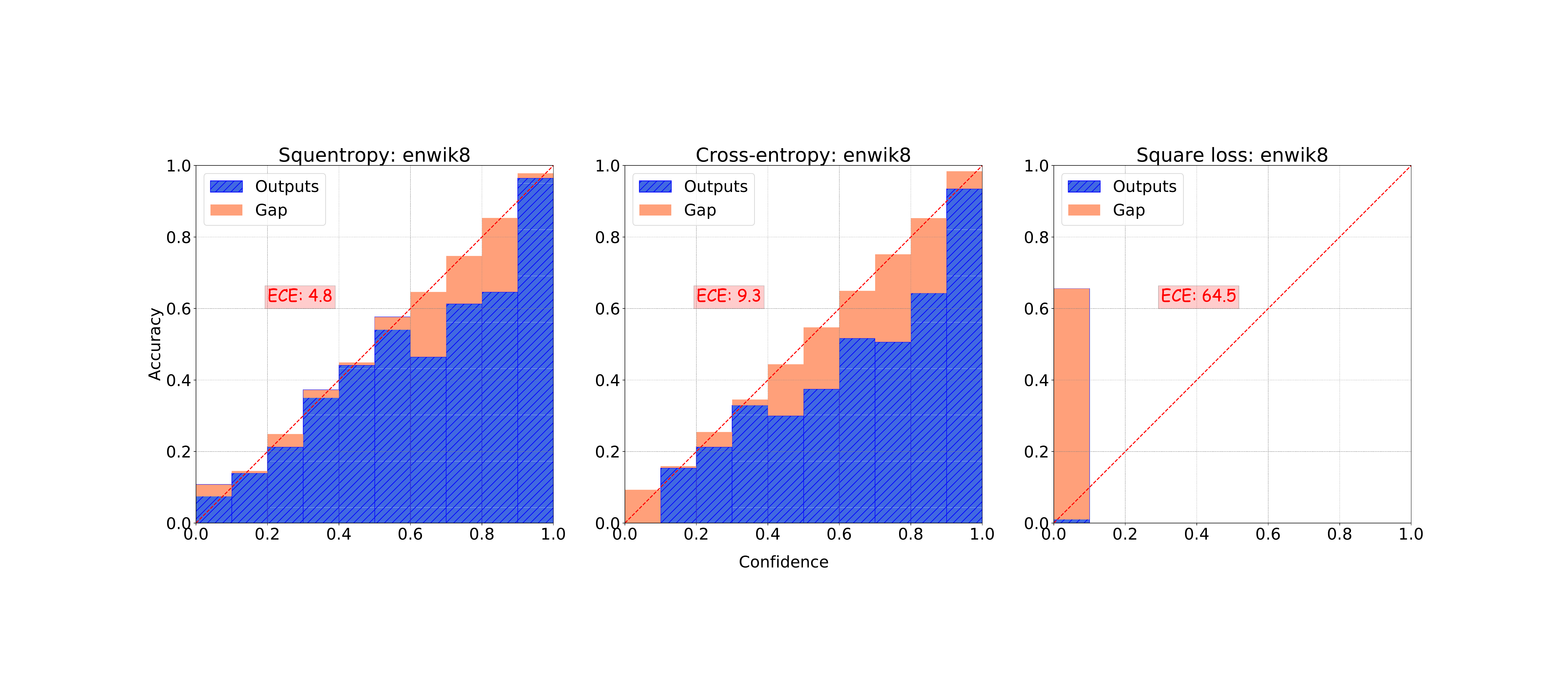}} 
\vspace{-15mm}
  \caption{Reliability diagrams for a Transformer-XL
on enwik8. \textit{Left:} squentropy, \textit{middle:} cross-entropy, \textit{right:} square loss.}
\end{figure*}
\begin{figure*}[!htbp]
  % \centering
\centerline{\includegraphics[width=1\columnwidth]{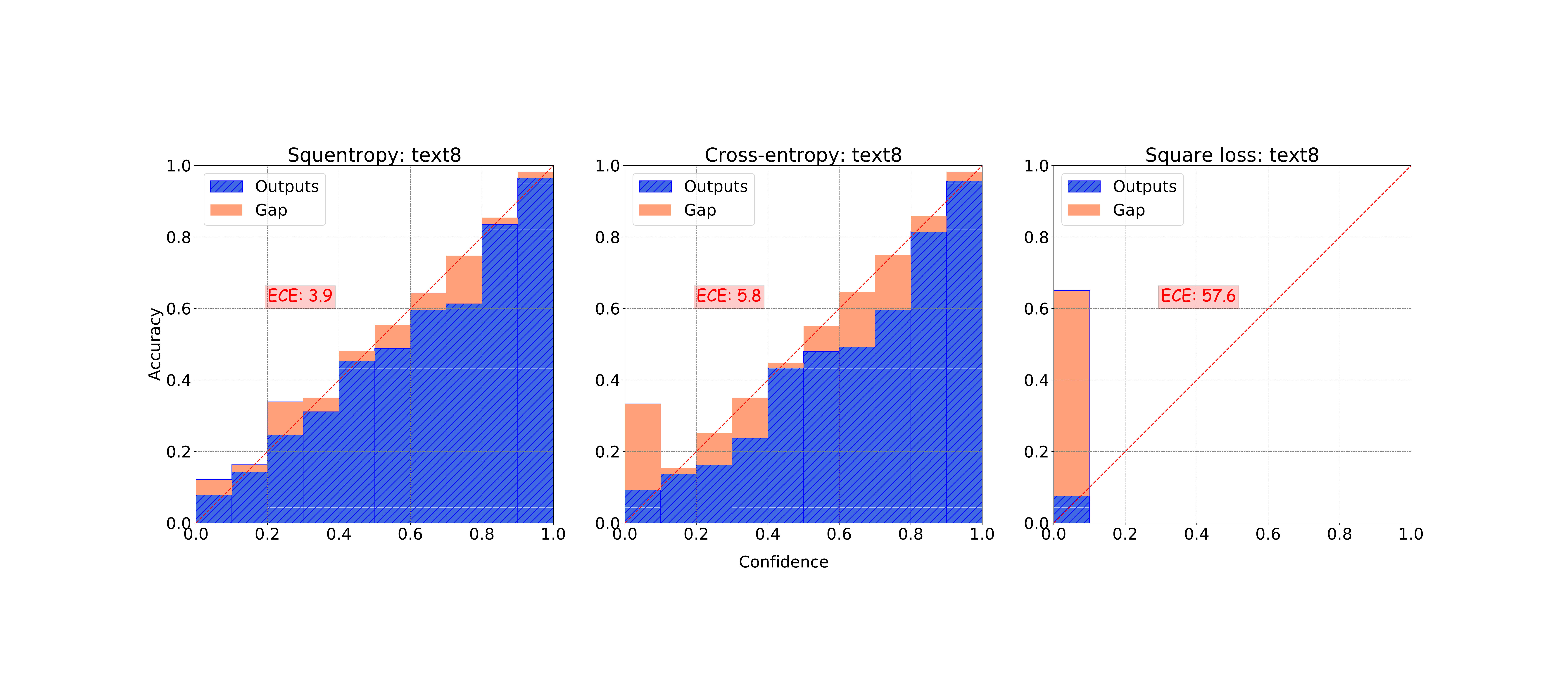}} 
\vspace{-15mm}
  \caption{Reliability diagrams for a Transformer-XL
on text8. \textit{Left:} squentropy, \textit{middle:} cross-entropy, \textit{right:} square loss.}
\end{figure*}
\begin{figure*}[!htbp]
  % \centering
\centerline{\includegraphics[width=1\columnwidth]{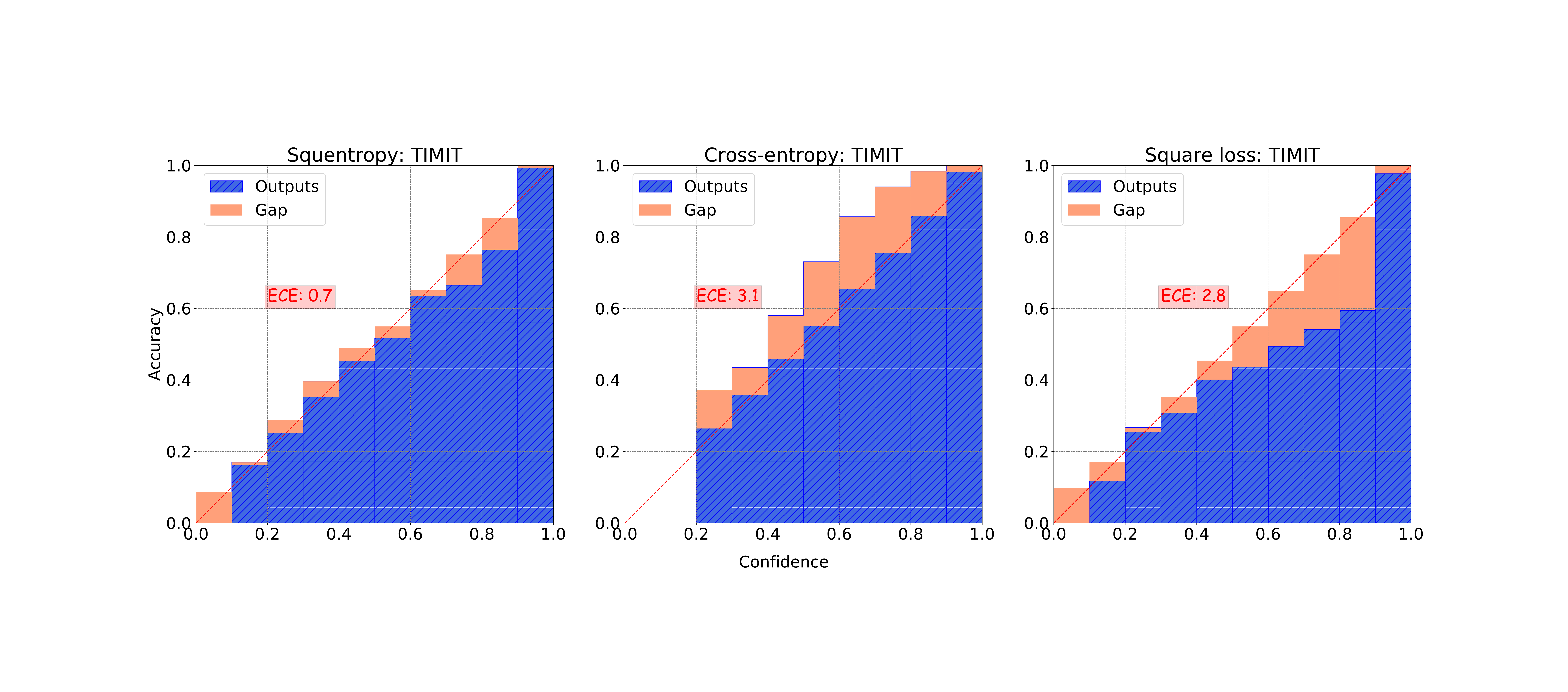}} 
\vspace{-15mm}
  \caption{Reliability diagrams for a Attention+CTC model
on TIMIT. \textit{Left:} squentropy, \textit{middle:} cross-entropy, \textit{right:} square loss.}
\end{figure*}

\begin{figure*}[!htbp]
  % \centering
\centerline{\includegraphics[width=1\columnwidth]{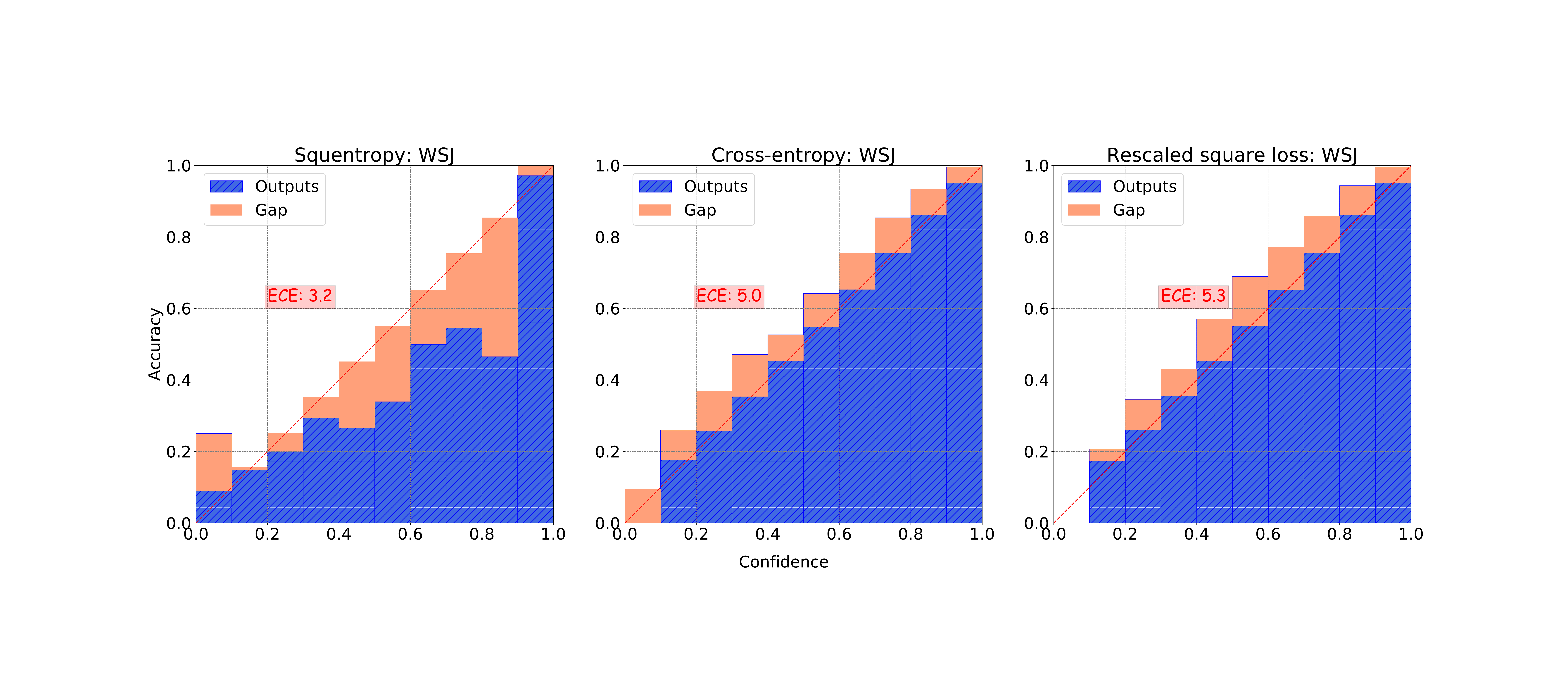}} 
\vspace{-15mm}
  \caption{Reliability diagrams for a VGG+BLSTMP model
on WSJ. \textit{Left:} squentropy, \textit{middle:} cross-entropy, \textit{right:} scaled square loss.}
\end{figure*}
\begin{figure*}[!htbp]
  % \centering
\centerline{\includegraphics[width=1\columnwidth]{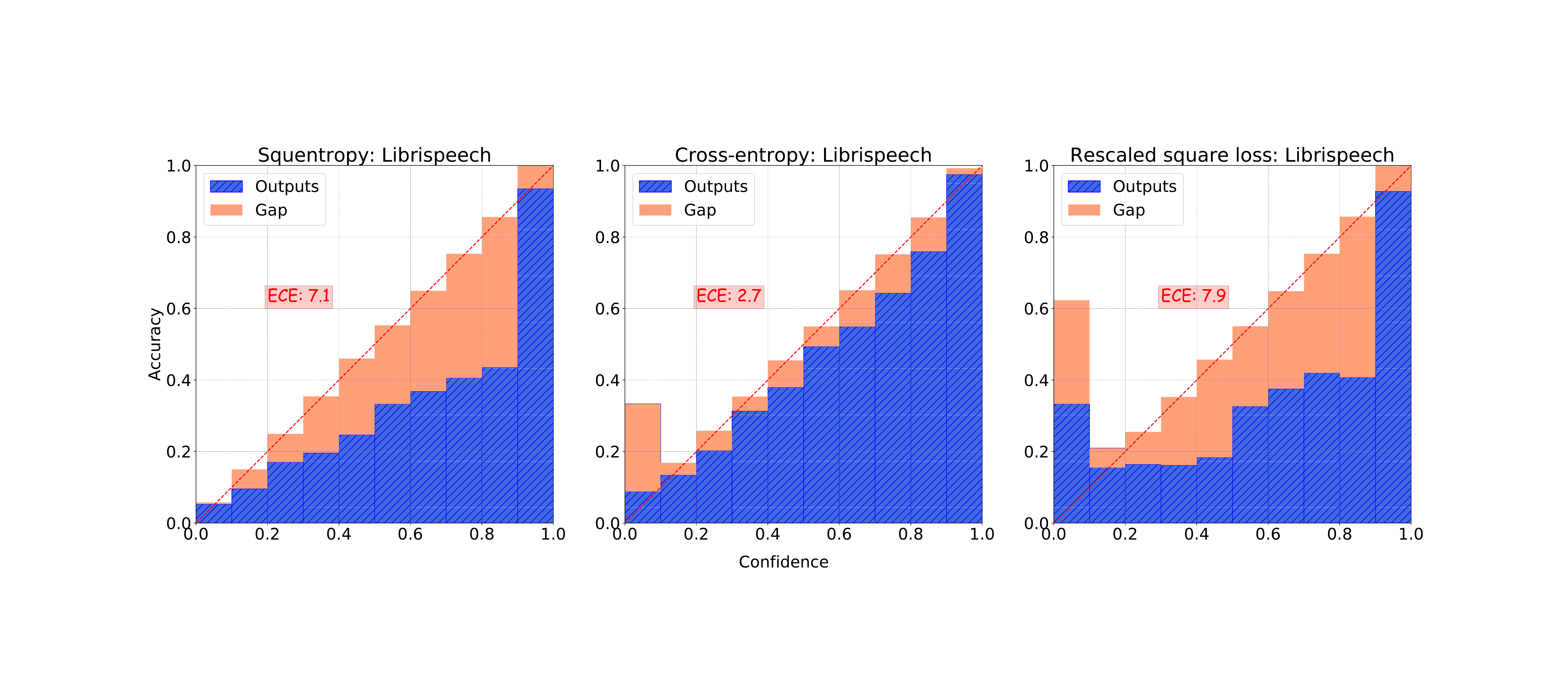}} 
\vspace{-15mm}
  \caption{Reliability diagrams for a VGG+BLSTM model
on Librispeech. \textit{Left:} squentropy, \textit{middle:} cross-entropy, \textit{right:} scaled square loss.}
\end{figure*}
\begin{figure*}[!htbp]
  % \centering
\centerline{\includegraphics[width=1\columnwidth]{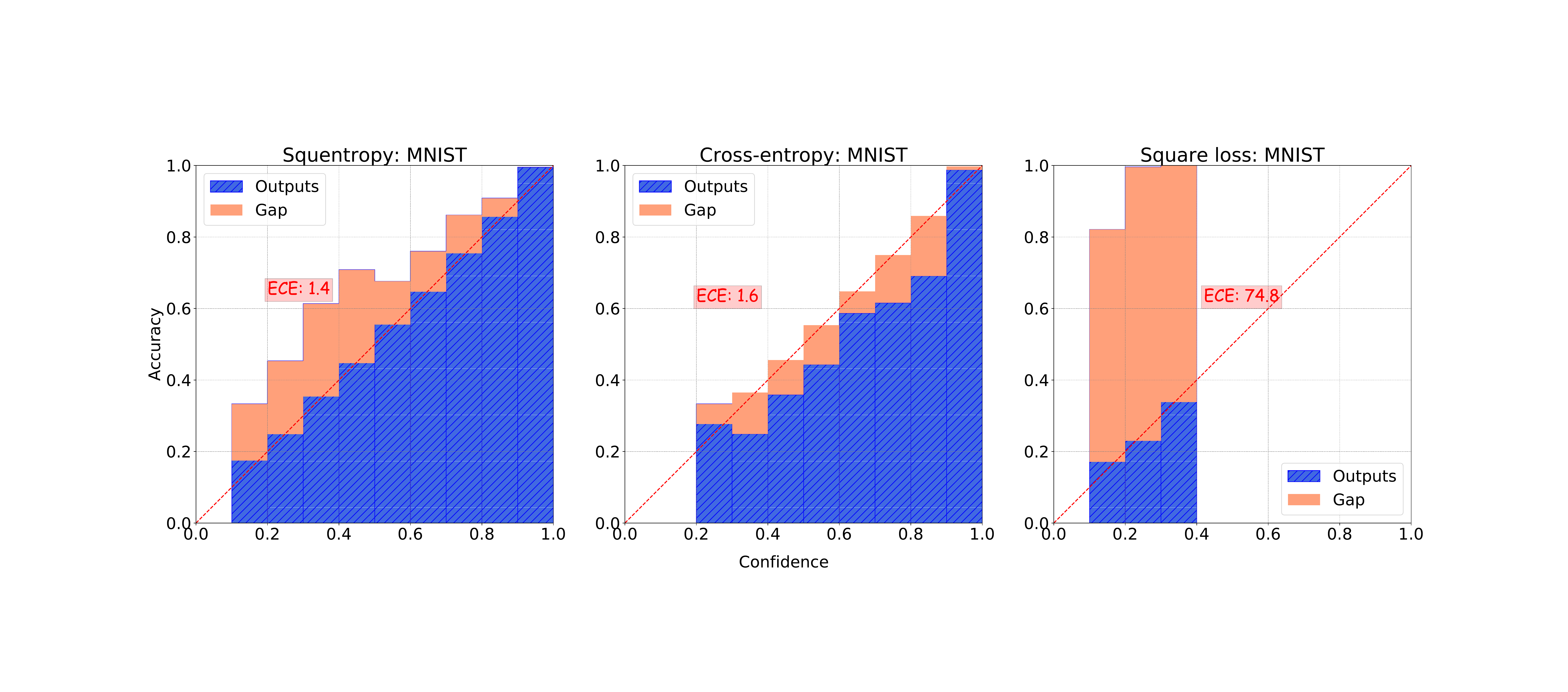}} 
\vspace{-15mm}
  \caption{Reliability diagrams for a TCN
on MNIST. \textit{Left:} squentropy, \textit{middle:} cross-entropy, \textit{right:} square loss.}
\end{figure*}

\begin{figure*}[!htbp]
  % \centering
\centerline{\includegraphics[width=1\columnwidth]{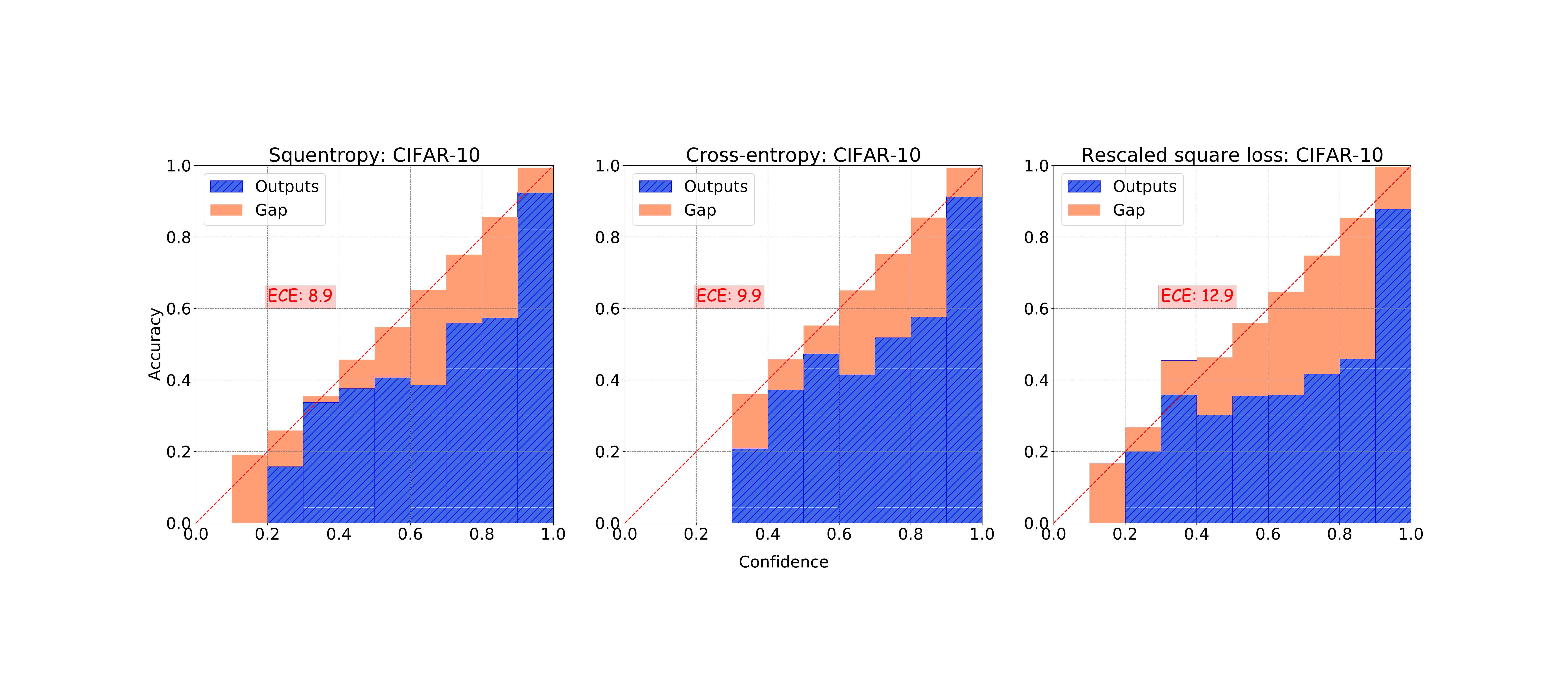}} 
\vspace{-15mm}
  \caption{Reliability diagrams for a Resnet18
on CIFAR-10.\textit{Left:} squentropy, \textit{middle:} cross-entropy, \textit{right:} scaled square loss.}
\end{figure*}

\begin{figure*}[!htbp]
  % \centering
\centerline{\includegraphics[width=1\columnwidth]{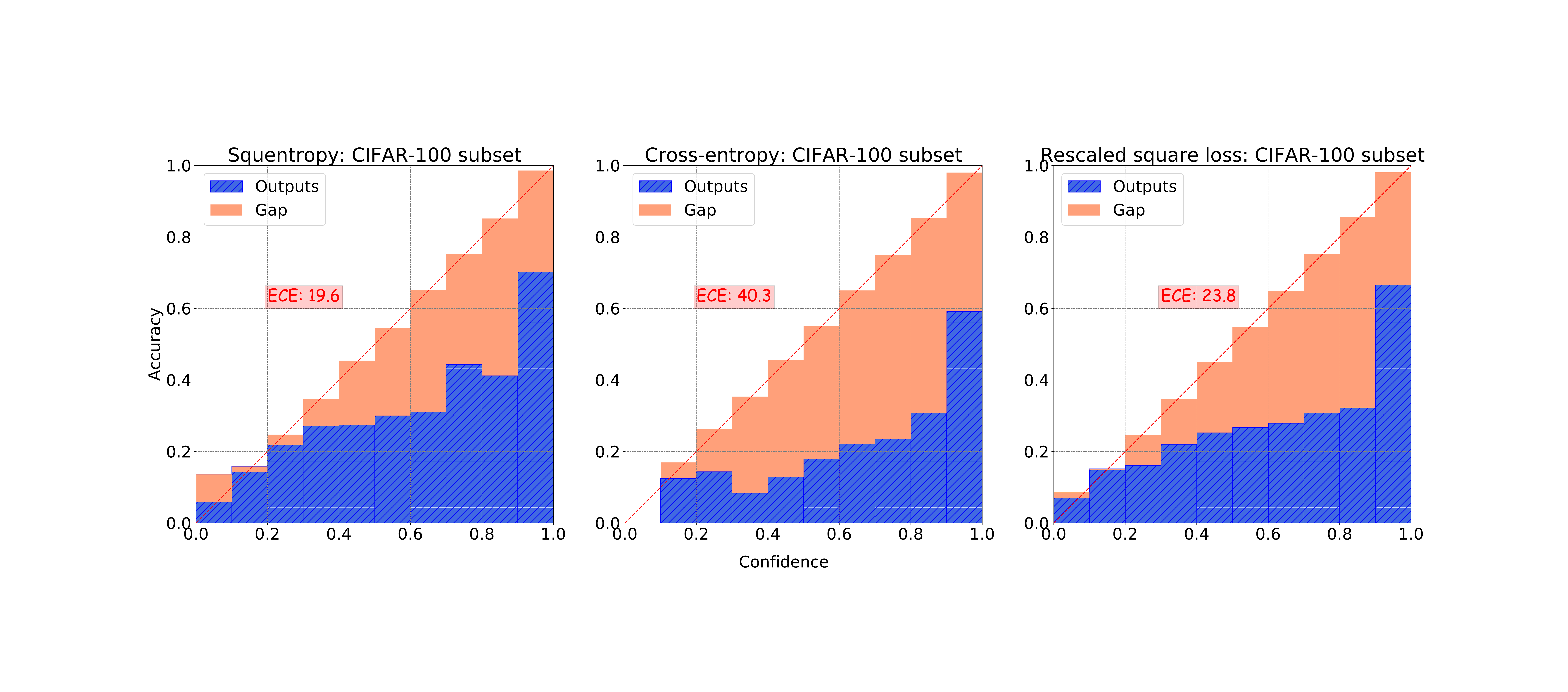}} 
\vspace{-15mm}
  \caption{Reliability diagrams for a Wide Resnet
on CIFAR-100 subset. \textit{Left:} squentropy, \textit{middle:} cross-entropy, \textit{right:} scaled square loss.}
\end{figure*}

\begin{figure*}[!htbp]
  % \centering
\centerline{\includegraphics[width=1\columnwidth]{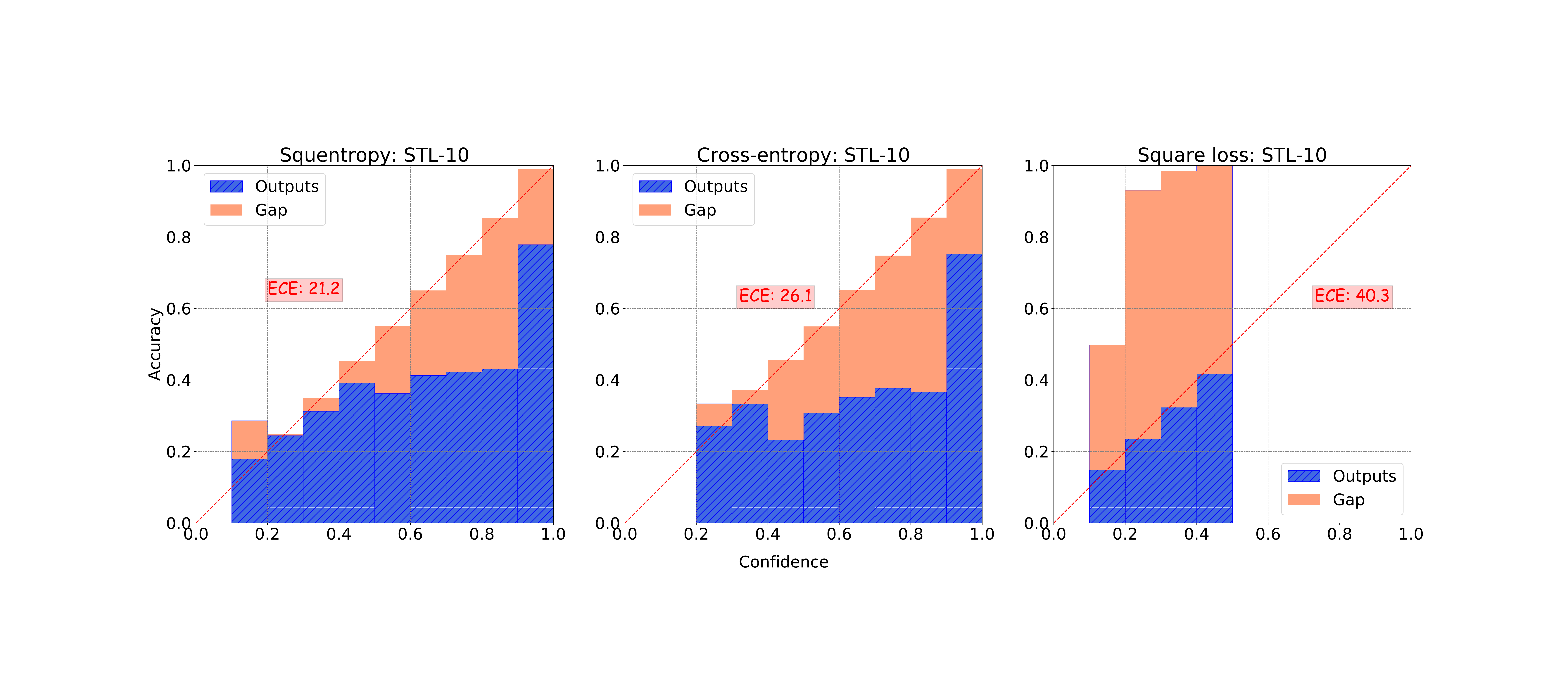}} 
\vspace{-15mm}
  \caption{Reliability diagrams for a Resnet18
on STL10. \textit{Left:} squentropy, \textit{middle:} cross-entropy, \textit{right:} square loss.}
\end{figure*}
\begin{figure*}[!htbp]
  % \centering
\centerline{\includegraphics[width=1\columnwidth]{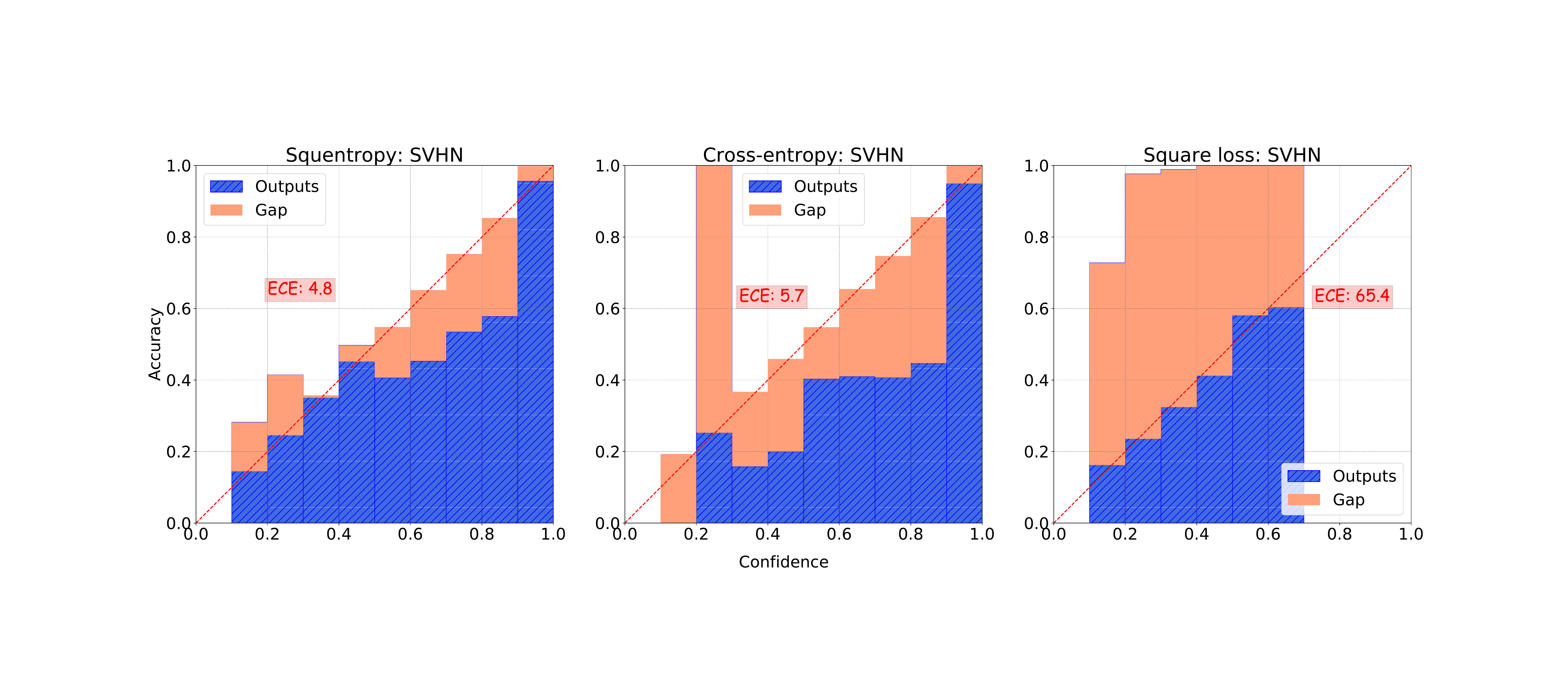}} 
\vspace{-15mm}
  \caption{Reliability diagrams for a VGG
on SVHN. \textit{Left:} squentropy, \textit{middle:} cross-entropy, \textit{right:} square loss.}
\end{figure*}

%% file: 121_table_results.tex
\begin{table}[!htbp]
\centering
\vspace{-3mm}
\caption{Test accuracy (Acc)/ECE for 121 tabular datasets}
%\begin{threeparttable}
\label{121t1}
\begin{tabular}{|c|cc|cc|cc|}
\hline
\multirow{2}{*}{Dataset} & \multicolumn{2}{c|}{Squentropy}      & \multicolumn{2}{c|}{Cross-entropy} & \multicolumn{2}{c|}{Rescaled square} \\ \cline{2-7} 
                         & \multicolumn{1}{c}{Acc}       & ECE & \multicolumn{1}{c}{Acc}   & ECE   & \multicolumn{1}{c}{Acc}  & ECE  \\ \hline
abalone & 66.0 & \textbf{3.9} & 67.9 & 14.1 & \textbf{68.3} & 13.8 \\ \hline 
acute-inflammation & \textbf{96.4} & \textbf{3.8} & 91.3 & 4.7 & 95.8 & 4.3 \\ \hline 
acute-nephritis & \textbf{100.0} & \textbf{1.7} & \textbf{100.0} & 3.0 & \textbf{100.0} & 4.8 \\ \hline 
adult & 84.0 & \textbf{4.7} & \textbf{85.1} & 5.7 & \textbf{85.1} & 10.7 \\ \hline 
annealing & 94.3 & \textbf{3.6} & 93.3 & 3.9 & \textbf{94.4} & 4.1 \\ \hline 
arrhythmia & 68.1 & 21.4 & 67.0 & 24.5 & \textbf{68.4} & \textbf{17.3} \\ \hline 
audiology-std & 72.6 & 26.3 & \textbf{73.0} & \textbf{26.0} & 70.3 & 29.7 \\ \hline 
balance-scale & \textbf{97.2} & 5.0 & 96.3 & \textbf{3.8} & 96.5 & 4.0 \\ \hline 
balloons & \textbf{95.6} & 23.7 & 95.4 & \textbf{21.4} & 90.0 & 25.8 \\ \hline 
bank & \textbf{89.9} & \textbf{4.1} & 89.8 & 7.7 & 89.2 & 10.6 \\ \hline 
blood & 81.6 & 11.7 & \textbf{81.9} & \textbf{7.0} & 81.7 & 16.6 \\ \hline 
breast-cancer & \textbf{76.8} & 26.6 & 75.6 & \textbf{24.5} & 75.5 & 27.5 \\ \hline 
breast-cancer-wisc & \textbf{98.0} & 4.8 & 97.6 & 4.9 & 97.1 & \textbf{4.5} \\ \hline 
breast-cancer-wisc-diag & \textbf{99.5} & 3.5 & 99.2 & 3.4 & 98.8 &\textbf{ 2.2} \\ \hline 
breast-cancer-wisc-prog & \textbf{89.6} & 16.3 & 87.9 & \textbf{15.7} & 89.0 & 19.3 \\ \hline 
breast-tissue & 83.3 & 17.9 & \textbf{84.1} & \textbf{17.1} & 83.6 & 19.2 \\ \hline 
car & \textbf{100.0} & \textbf{0.4} & \textbf{100.0} & \textbf{0.4} & \textbf{100.0} & 2.3 \\ \hline 
cardiotocography-10clases & 87.7 & 6.3 & \textbf{87.8} & 7.3 & 87.2 & \textbf{3.9} \\ \hline 
cardiotocography-3clases & 94.8 & \textbf{4.0} & \textbf{94.9} & 5.5 & 94.7 & 4.6 \\ \hline 
chess-krvk & 86.6 & 3.7 & \textbf{87.8} & \textbf{1.8} & 86.1 & 16.0 \\ \hline 
chess-krvkp & \textbf{99.8} & \textbf{0.4} & 99.7 & 0.5 & \textbf{99.8} & 0.7 \\ \hline 
congressional-voting & \textbf{65.9} & 9.7 & 65.7 & \textbf{9.2} & 65.1 & 18.3 \\ \hline 
conn-bench-sonar-mines-rocks & 90.6 & 11.7 & \textbf{91.4} & \textbf{10.2} & \textbf{91.4} & 13.1 \\ \hline 
conn-bench-vowel-deterding & \textbf{99.6} & 2.7 & 99.1 & \textbf{1.9} & 98.9 & 9.3 \\ \hline 
connect-4 & 89.4 & \textbf{3.3} & 89.0 & 5.1 & \textbf{89.7} & 6.6 \\ \hline 
contrac & 57.7 & \textbf{13.6} & 58.6 & 30.4 & \textbf{58.0} & 35.3 \\ \hline 
credit-approval & \textbf{89.1} & \textbf{9.4} & 88.4 & 11.7 & 88.6 & 14.3 \\ \hline 
cylinder-bands & 81.9 & \textbf{13.1} &\textbf{ 84.1} & 14.8 & 83.9 & 17.8 \\ \hline 
dermatology & \textbf{97.6} & 4.7 & 97.2 & 4.4 & 97.3 & \textbf{3.5} \\ \hline 
echocardiogram & \textbf{85.0} & \textbf{17.6} & 84.9 & 17.8 & 84.4 & 19.3 \\ \hline 
ecoli & \textbf{88.2} & 12.4 & 88.1 & \textbf{7.8} & \textbf{88.2} & 9.8 \\ \hline 
energy-y1 & 97.3 & 4.2 & \textbf{97.5} & 4.2 & 97.3 & \textbf{3.0} \\ \hline 
energy-y2 & \textbf{97.2} & 4.4 & 96.7 & 5.3 & 96.4 & \textbf{4.0} \\ \hline 
fertility & \textbf{94.6} & 23.3 & 93.4 & 26.3 & 90.0 & \textbf{19.4} \\ \hline 
flags & \textbf{60.1} & 27.1 & 58.4 & 31.3 & 57.4 & \textbf{20.6} \\ \hline 
glass & \textbf{78.7} & \textbf{18.7} & 78.1 & 25.4 & 78.1 & 20.8 \\ \hline 
haberman-survival & \textbf{80.3} & \textbf{12.2} & 79.8 & 17.9 & 79.7 & 22.9 \\ \hline 
hayes-roth & \textbf{85.8} & \textbf{5.4} & 84.1 & 11.4 & 83.7 & 13.6 \\ \hline 
heart-cleveland & 62.5 & 32.0 & 62.1 & 32.5 & \textbf{64.6} & \textbf{25.7} \\ \hline 
heart-hungarian & 85.3 & 17.6 & 84.8 & \textbf{16.0} & \textbf{85.4} & 19.2 \\ \hline 
heart-switzerland & 43.8 & 50.4 & 43.6 & 47.7 & \textbf{46.4} &\textbf{ 44.4} \\ \hline 
heart-va & 42.1 & \textbf{46.0} & \textbf{46.4} & 54.1 & 42.0 & 47.9 \\ \hline 
hepatitis & 77.4 & 18.1 & 75.3 & 20.6 & \textbf{84.5} & \textbf{14.8} \\ \hline 
hill-valley & 66.0 & \textbf{14.2} & \textbf{71.9} & 36.3 & 71.1 & 40.2 \\ \hline 
horse-colic & 85.9 & \textbf{13.3} & 84.4 & 13.9 & \textbf{86.0} & 14.2 \\ \hline 

\end{tabular}
\end{table}

\begin{table}[ht]
\centering
\vspace{-3mm}
\caption{Test accuracy (Acc)/ECE for 121 tabular datasets}
%\begin{threeparttable}
\label{121t2}
\begin{tabular}{|c|cc|cc|cc|}
\hline
\multirow{2}{*}{Dataset} & \multicolumn{2}{c|}{Squentropy}      & \multicolumn{2}{c|}{Cross-entropy} & \multicolumn{2}{c|}{Rescaled square} \\ \cline{2-7} 
                         & \multicolumn{1}{c}{Acc}       & ECE & \multicolumn{1}{c}{Acc}   & ECE   & \multicolumn{1}{c}{Acc}  & ECE  \\ \hline
ilpd-indian-liver & \textbf{77.2} & \textbf{12.7} & 75.3 & 22.9 & 75.9 & 25.7 \\ \hline 
image-segmentation & \textbf{96.8} & \textbf{5.4} & 94.7 & 6.5 & 94.8 & 6.9 \\ \hline 
ionosphere & \textbf{98.3} & 7.0 & 98.2 &\textbf{ 5.5} & 97.2 & 6.2 \\ \hline 
iris & 97.2 & 3.9 & 97.1 & 4.3 & \textbf{98.0} & \textbf{2.9} \\ \hline 
led-display & 75.2 & 10.7 & \textbf{75.3} & \textbf{5.2} & 74.9 & 8.3 \\ \hline 
lenses & 76.6 & 20.8 & 68.4 & 21.5 & \textbf{80.0} & \textbf{17.8} \\ \hline 
letter & \textbf{98.8} & \textbf{1.1} & 98.6 & 1.2 & 98.4 & 16.6 \\ \hline 
libras & \textbf{93.1} & \textbf{4.9} & 92.9 & 5.7 & 92.5 & 12.9 \\ \hline 
low-res-spect & \textbf{95.9} & \textbf{4.0} & 95.2 & 5.0 & 94.2 & 7.7 \\ \hline 
lung-cancer & 60.6 & \textbf{27.1} & 54.7 & 40.4 & \textbf{62.9} & 40.4 \\ \hline 
lymphography & 89.2 & \textbf{6.4} & 87.1 & 7.1 & \textbf{91.3} & 8.0 \\ \hline 
magic & 88.3 & 5.5 & 89.1 & \textbf{5.3} & \textbf{89.2} & 8.3 \\ \hline 
mammographic & 81.9 & 9.4 & 83.3 & \textbf{8.0} & \textbf{83.4} & 14.6 \\ \hline 
miniboone & \textbf{81.7} & \textbf{20.3} & 81.5 & \textbf{20.3} & 77.9 & 27.2 \\ \hline 
molec-biol-promoter & \textbf{87.9} & 11.4 & 78.6 & \textbf{9.9} & 85.5 & 14.8 \\ \hline 
molec-biol-splice & \textbf{87.9} & \textbf{7.1} & 84.2 & 10.8 & 87.2 & 8.2 \\ \hline 
monks-1 & 85.4 & 14.0 & 83.6 & \textbf{13.5} & \textbf{87.2} & 14.6 \\ \hline 
monks-2 & 72.9 & \textbf{6.9} & 89.9 & 12.8 & \textbf{95.9} & 14.5 \\ \hline 
monks-3 & \textbf{93.4} & \textbf{6.7} & 91.6 & 7.6 & 92.0 & 9.4 \\ \hline 
mushroom & \textbf{100.0} & \textbf{0.0} & \textbf{100.0} & \textbf{0.0} & \textbf{100.0} & 0.7 \\ \hline 
musk-1 & 95.2 & \textbf{3.3} & 94.6 & 6.3 & \textbf{95.6} & 4.9 \\ \hline 
musk-2 & \textbf{100.0} & 1.1 & \textbf{100.0} & 0.2 & \textbf{100.0} & \textbf{0.7} \\ \hline 
nursery & \textbf{100.0} & 0.1 & \textbf{100.0} & \textbf{0.0} & \textbf{100.0} & 2.7 \\ \hline 
oocytes\_merluccius\_nucleus\_4d & 85.1 & \textbf{3.5} & 86.6 & 10.4 & \textbf{87.1} & 14.4 \\ \hline 
oocytes\_merluccius\_states\_2f & \textbf{95.4} & \textbf{3.2} & 95.2 & 6.3 & 95.2 & 4.6 \\ \hline 
oocytes\_trisopterus\_nucleus\_2f & 89.0 & \textbf{5.4} & \textbf{89.7} & 9.2 & 89.0 & 10.7 \\ \hline 
oocytes\_trisopterus\_states\_5b & 97.1 & 4.3 & 97.1 & 4.6 & \textbf{97.3} & \textbf{3.7} \\ \hline 
optical & \textbf{99.6} & \textbf{0.9} & 99.3 & 1.2 & 99.0 & 6.6 \\ \hline 
ozone & \textbf{97.7} & 4.5 & 97.5 & 3.9 & 97.1 & \textbf{3.1} \\ \hline 
page-blocks & \textbf{97.7} & 2.2 & 97.5 & 2.3 & 97.1 & \textbf{1.8} \\ \hline 
parkinsons & 97.0 & \textbf{2.9} & \textbf{97.9} & 5.8 & 97.4 & 4.2 \\ \hline 
pendigits & 99.8 & \textbf{0.2} & 99.8 & 0.3 & \textbf{99.9} & 6.1 \\ \hline 
pima & \textbf{79.9} & \textbf{20.8} & 78.0 & 22.4 & 77.4 & 24.6 \\ \hline 
pittsburg-bridges-MATERIAL & 79.7 & 15.4 & 80.4 & 15.8 & \textbf{89.1} & \textbf{14.0} \\ \hline 
pittsburg-bridges-REL-L & 68.2 & 30.2 & 66.2 & \textbf{28.3} & \textbf{73.3} & 36.5 \\ \hline 
pittsburg-bridges-SPAN & 73.2 & 31.0 & 69.9 & \textbf{30.8} & \textbf{73.7} & 34.6 \\ \hline 
pittsburg-bridges-T-OR-D & \textbf{90.1} & 19.3 & 90.0 & 24.5 & 89.5 & \textbf{17.9} \\ \hline 
pittsburg-bridges-TYPE & 63.4 & 34.8 & 63.3 & \textbf{33.0} & \textbf{66.7} & 39.7 \\ \hline 
planning & \textbf{77.9} & \textbf{23.4} & 75.6 & 33.5 & 76.8 & 31.7 \\ \hline 
plant-margin & \textbf{85.1} & \textbf{4.4} & 84.0 & 5.9 & 82.9 & 55.7 \\ \hline 
plant-shape & \textbf{74.3} & \textbf{7.8} & 73.9 & 13.9 & 70.6 & 49.1 \\ \hline 
plant-texture & \textbf{85.3} & \textbf{3.1} & 84.3 & 6.2 & 82.6 & 52.8 \\ \hline 
post-operative & \textbf{73.9} & 35.2 & 70.4 & \textbf{34.7} & 62.2 & 35.3 \\ \hline 
primary-tumor & \textbf{50.0} & 27.7 & 49.8 & 38.5 & 48.5 & \textbf{24.0} \\ \hline 
ringnorm & \textbf{98.6} & 1.7 & 98.5 & 2.1 & 98.1 & \textbf{1.5} \\ \hline 

\end{tabular}
\end{table}

\begin{table}[ht]
\centering
\vspace{-3mm}
\caption{Test accuracy (Acc)/ECE for 121 tabular datasets}
\label{121t3}

\begin{tabular}{|c|cc|cc|cc|}
\hline
\multirow{2}{*}{Dataset} & \multicolumn{2}{c|}{Squentropy}      & \multicolumn{2}{c|}{Cross-entropy} & \multicolumn{2}{c|}{Rescaled square} \\ \cline{2-7} 
                         & \multicolumn{1}{c}{Acc}       & ECE & \multicolumn{1}{c}{Acc}   & ECE   & \multicolumn{1}{c}{Acc}  & ECE  \\ \hline
seeds & \textbf{100.0} & \textbf{4.0} & 99.0 & 6.3 & 98.6 & 5.9 \\ \hline 
semeion & \textbf{95.4} & \textbf{2.3} & 94.9 & 3.5 & 94.8 & 10.7 \\ \hline 
soybean & \textbf{92.2} & \textbf{3.4} & 90.8 & 3.9 & 91.3 & 17.5 \\ \hline 
spambase & \textbf{95.6} & \textbf{4.1} & 95.4 & 4.6 & 95.0 & 4.9 \\ \hline 
spect & \textbf{76.8} & \textbf{37.3} & 75.4 & 41.7 & 76.2 & 42.2 \\ \hline 
spectf & 79.3 & 18.2 & 82.9 & \textbf{17.1} & \textbf{83.8} & 21.8 \\ \hline 
statlog-australian-credit & 61.2 & \textbf{24.1} & 64.5 & 34.0 & \textbf{66.4} & 34.1 \\ \hline 
statlog-german-credit & \textbf{80.3} & \textbf{15.7} & 79.1 & 21.2 & 79.6 & 23.1 \\ \hline 
statlog-heart & \textbf{86.9} & 18.0 & 86.4 & \textbf{15.4} & 85.9 & 18.8 \\ \hline 
statlog-image & \textbf{99.3} & \textbf{1.4} & 99.1 & \textbf{1.4} & 98.8 & 4.1 \\ \hline 
statlog-landsat & \textbf{93.3} & 5.8 & 93.0 & 6.8 & 92.7 & \textbf{2.7} \\ \hline 
statlog-shuttle & \textbf{99.8} & \textbf{0.5} & \textbf{99.8} & \textbf{0.5} & \textbf{99.8} & 3.7 \\ \hline 
statlog-vehicle & \textbf{87.5} & \textbf{7.8} & 86.9 & 9.7 & 86.8 & 11.9 \\ \hline 
steel-plates & \textbf{78.8} & \textbf{9.7} & 78.5 & 14.4 & 78.7 & 10.2 \\ \hline 
synthetic-control &\textbf{ 99.1} & 2.1 & \textbf{99.1} & \textbf{2.0} & 98.7 & 4.9 \\ \hline 
teaching & \textbf{65.1} & \textbf{29.9} & 63.0 & 30.5 & 63.9 & 37.2 \\ \hline 
thyroid & \textbf{98.5} & 2.0 & 97.6 & 2.6 & 97.9 & \textbf{1.4} \\ \hline 
tic-tac-toe & \textbf{99.8} & 0.3 & \textbf{99.8} & \textbf{0.2} & \textbf{99.8} & 0.6 \\ \hline 
titanic & 78.6 & 12.6 & 78.4 & \textbf{4.2} & \textbf{78.9} & 13.6 \\ \hline 
trains &\textbf{ 100.0} & 34.1 & 90.4 & \textbf{27.2} & 80.0 & 53.0 \\ \hline 
twonorm & \textbf{98.2} & \textbf{2.0} & 98.1 & 2.6 & 97.7 & \textbf{2.0} \\ \hline 
vertebral-column-2clases & 91.2 & \textbf{8.5} & 91.1 & 8.6 & \textbf{91.3} & 13.1 \\ \hline 
vertebral-column-3clases & \textbf{88.0} & 15.4 & 86.6 & \textbf{15.1} & 87.1 & 16.1 \\ \hline 
wall-following & \textbf{96.1} & 2.2 & 95.8 & 3.2 & 95.7 & \textbf{1.7} \\ \hline 
waveform & 86.5 & \textbf{9.0} & 86.8 & 11.2 & \textbf{86.9} & 12.0 \\ \hline 
waveform-noise & 85.4 & \textbf{9.4} & 85.4 & 11.9 & \textbf{86.1} & 14.1 \\ \hline 
wine & \textbf{100.0} & 3.3 & \textbf{100.0} & 3.0 & \textbf{100.0} & \textbf{2.9} \\ \hline 
wine-quality-red & 68.8 & 23.2 & 68.9 & 26.6 & \textbf{69.3} & \textbf{19.7} \\ \hline 
wine-quality-white & 65.0 & 19.9 & \textbf{65.9} & 25.3 & 65.5 & \textbf{17.2} \\ \hline 
yeast & 63.3 & 21.4 & 63.1 & 29.9 & \textbf{63.5} & \textbf{18.8} \\ \hline 
zoo & \textbf{92.0} & 4.8 & 91.9 & \textbf{3.9} & 91.4 & 9.1 \\ \hline 

\end{tabular}
\end{table}

%% file: icml2023.bbl
\begin{thebibliography}{35}
\providecommand{\natexlab}[1]{#1}
\providecommand{\url}[1]{\texttt{#1}}
\expandafter\ifx\csname urlstyle\endcsname\relax
  \providecommand{\doi}[1]{doi: #1}\else
  \providecommand{\doi}{doi: \begingroup \urlstyle{rm}\Url}\fi

\bibitem[Bai et~al.(2018)Bai, Kolter, and Koltun]{bai2018empirical}
Bai, S., Kolter, J.~Z., and Koltun, V.
\newblock An empirical evaluation of generic convolutional and recurrent
  networks for sequence modeling.
\newblock \emph{arXiv preprint arXiv:1803.01271}, 2018.

\bibitem[Chen et~al.(2016)Chen, Zhu, Ling, Wei, Jiang, and
  Inkpen]{chen2016enhanced}
Chen, Q., Zhu, X., Ling, Z., Wei, S., Jiang, H., and Inkpen, D.
\newblock Enhanced lstm for natural language inference.
\newblock \emph{arXiv preprint arXiv:1609.06038}, 2016.

\bibitem[Coates et~al.(2011)Coates, Ng, and Lee]{coates2011analysis}
Coates, A., Ng, A., and Lee, H.
\newblock An analysis of single-layer networks in unsupervised feature
  learning.
\newblock In \emph{Proceedings of the fourteenth international conference on
  artificial intelligence and statistics}, pp.\  215--223. JMLR Workshop and
  Conference Proceedings, 2011.

\bibitem[Dai et~al.(2019)Dai, Yang, Yang, Carbonell, Le, and
  Salakhutdinov]{dai2019transformer}
Dai, Z., Yang, Z., Yang, Y., Carbonell, J., Le, Q.~V., and Salakhutdinov, R.
\newblock Transformer-xl: Attentive language models beyond a fixed-length
  context.
\newblock \emph{arXiv preprint arXiv:1901.02860}, 2019.

\bibitem[DeGroot \& Fienberg(1983)DeGroot and Fienberg]{degroot1983comparison}
DeGroot, M.~H. and Fienberg, S.~E.
\newblock The comparison and evaluation of forecasters.
\newblock \emph{Journal of the Royal Statistical Society: Series D (The
  Statistician)}, 32\penalty0 (1-2):\penalty0 12--22, 1983.

\bibitem[Demirkaya et~al.(2020)Demirkaya, Chen, and
  Oymak]{demirkaya2020exploring}
Demirkaya, A., Chen, J., and Oymak, S.
\newblock Exploring the role of loss functions in multiclass classification.
\newblock In \emph{2020 54th annual conference on information sciences and
  systems (ciss)}, pp.\  1--5. IEEE, 2020.

\bibitem[Devlin et~al.(2018)Devlin, Chang, Lee, and Toutanova]{devlin2018bert}
Devlin, J., Chang, M.-W., Lee, K., and Toutanova, K.
\newblock Bert: Pre-training of deep bidirectional transformers for language
  understanding.
\newblock \emph{arXiv preprint arXiv:1810.04805}, 2018.

\bibitem[Elsayed et~al.(2018)Elsayed, Krishnan, Mobahi, Regan, and
  Bengio]{elsayed2018large}
Elsayed, G., Krishnan, D., Mobahi, H., Regan, K., and Bengio, S.
\newblock Large margin deep networks for classification.
\newblock \emph{Advances in neural information processing systems}, 31, 2018.

\bibitem[Fern{\'a}ndez-Delgado et~al.(2014)Fern{\'a}ndez-Delgado, Cernadas,
  Barro, and Amorim]{fernandez2014we}
Fern{\'a}ndez-Delgado, M., Cernadas, E., Barro, S., and Amorim, D.
\newblock Do we need hundreds of classifiers to solve real world classification
  problems?
\newblock \emph{The journal of machine learning research}, 15\penalty0
  (1):\penalty0 3133--3181, 2014.

\bibitem[Guo et~al.(2017)Guo, Pleiss, Sun, and Weinberger]{guo2017calibration}
Guo, C., Pleiss, G., Sun, Y., and Weinberger, K.~Q.
\newblock On calibration of modern neural networks.
\newblock In \emph{International conference on machine learning}, pp.\
  1321--1330. PMLR, 2017.

\bibitem[Han et~al.(2021)Han, Papyan, and Donoho]{han2021neural}
Han, X., Papyan, V., and Donoho, D.~L.
\newblock Neural collapse under mse loss: Proximity to and dynamics on the
  central path.
\newblock In \emph{International Conference on Learning Representations}, 2021.
\newblock arXiv preprint arXiv:2106.02073.

\bibitem[He \& Lin(2016)He and Lin]{he2016pairwise}
He, H. and Lin, J.
\newblock Pairwise word interaction modeling with deep neural networks for
  semantic similarity measurement.
\newblock In \emph{Proceedings of the 2016 conference of the north American
  chapter of the Association for Computational Linguistics: human language
  technologies}, pp.\  937--948, 2016.

\bibitem[He et~al.(2016)He, Zhang, Ren, and Sun]{he2016deep}
He, K., Zhang, X., Ren, S., and Sun, J.
\newblock Deep residual learning for image recognition.
\newblock In \emph{Proceedings of the IEEE conference on computer vision and
  pattern recognition}, pp.\  770--778, 2016.

\bibitem[Hui \& Belkin(2020)Hui and Belkin]{hui2020evaluation}
Hui, L. and Belkin, M.
\newblock Evaluation of neural architectures trained with square loss vs
  cross-entropy in classification tasks.
\newblock \emph{arXiv preprint arXiv:2006.07322}, 2020.

\bibitem[Kim et~al.(2017)Kim, Hori, and Watanabe]{kim2017joint}
Kim, S., Hori, T., and Watanabe, S.
\newblock Joint ctc-attention based end-to-end speech recognition using
  multi-task learning.
\newblock In \emph{2017 IEEE international conference on acoustics, speech and
  signal processing (ICASSP)}, pp.\  4835--4839. IEEE, 2017.

\bibitem[Krizhevsky et~al.(2009)Krizhevsky, Hinton,
  et~al.]{krizhevsky2009learning}
Krizhevsky, A., Hinton, G., et~al.
\newblock Learning multiple layers of features from tiny images.
\newblock 2009.

\bibitem[Lin et~al.(2017)Lin, Goyal, Girshick, He, and
  Doll{\'a}r]{lin2017focal}
Lin, T.-Y., Goyal, P., Girshick, R., He, K., and Doll{\'a}r, P.
\newblock Focal loss for dense object detection.
\newblock In \emph{Proceedings of the IEEE international conference on computer
  vision}, pp.\  2980--2988, 2017.

\bibitem[Liu et~al.(2022)Liu, Ben~Ayed, Galdran, and Dolz]{liu2022devil}
Liu, B., Ben~Ayed, I., Galdran, A., and Dolz, J.
\newblock The devil is in the margin: Margin-based label smoothing for network
  calibration.
\newblock In \emph{Proceedings of the IEEE/CVF Conference on Computer Vision
  and Pattern Recognition}, pp.\  80--88, 2022.

\bibitem[Moritz et~al.(2019)Moritz, Hori, and Le~Roux]{moritz2019triggered}
Moritz, N., Hori, T., and Le~Roux, J.
\newblock Triggered attention for end-to-end speech recognition.
\newblock In \emph{ICASSP 2019-2019 IEEE International Conference on Acoustics,
  Speech and Signal Processing (ICASSP)}, pp.\  5666--5670. IEEE, 2019.

\bibitem[Mukhoti et~al.(2020)Mukhoti, Kulharia, Sanyal, Golodetz, Torr, and
  Dokania]{mukhoti2020calibrating}
Mukhoti, J., Kulharia, V., Sanyal, A., Golodetz, S., Torr, P., and Dokania, P.
\newblock Calibrating deep neural networks using focal loss.
\newblock \emph{Advances in Neural Information Processing Systems},
  33:\penalty0 15288--15299, 2020.

\bibitem[M{\"u}ller et~al.(2019)M{\"u}ller, Kornblith, and
  Hinton]{muller2019does}
M{\"u}ller, R., Kornblith, S., and Hinton, G.~E.
\newblock When does label smoothing help?
\newblock \emph{Advances in neural information processing systems}, 32, 2019.

\bibitem[Naeini et~al.(2015)Naeini, Cooper, and
  Hauskrecht]{naeini2015obtaining}
Naeini, M.~P., Cooper, G., and Hauskrecht, M.
\newblock Obtaining well calibrated probabilities using bayesian binning.
\newblock In \emph{Proceedings of the AAAI conference on artificial
  intelligence}, volume~29, 2015.

\bibitem[Netzer et~al.(2011)Netzer, Wang, Coates, Bissacco, Wu, and
  Ng]{netzer2011reading}
Netzer, Y., Wang, T., Coates, A., Bissacco, A., Wu, B., and Ng, A.~Y.
\newblock Reading digits in natural images with unsupervised feature learning.
\newblock 2011.

\bibitem[Niculescu-Mizil \& Caruana(2005)Niculescu-Mizil and
  Caruana]{niculescu2005predicting}
Niculescu-Mizil, A. and Caruana, R.
\newblock Predicting good probabilities with supervised learning.
\newblock In \emph{Proceedings of the 22nd international conference on Machine
  learning}, pp.\  625--632, 2005.

\bibitem[Papyan et~al.(2020)Papyan, Han, and Donoho]{Pap20a}
Papyan, V., Han, X.~Y., and Donoho, D.~L.
\newblock Prevalence of neural collapse during the terminal phase of deep
  learning training.
\newblock \emph{Proceedings of the National Academy of Sciences}, 117\penalty0
  (40):\penalty0 24652--24663, 2020.
\newblock \doi{10.1073/pnas.2015509117}.
\newblock URL \url{https://www.pnas.org/doi/abs/10.1073/pnas.2015509117}.

\bibitem[Platt et~al.(1999)]{platt1999probabilistic}
Platt, J. et~al.
\newblock Probabilistic outputs for support vector machines and comparisons to
  regularized likelihood methods.
\newblock \emph{Advances in large margin classifiers}, 10\penalty0
  (3):\penalty0 61--74, 1999.

\bibitem[Que \& Belkin(2016)Que and Belkin]{que2016back}
Que, Q. and Belkin, M.
\newblock Back to the future: Radial basis function networks revisited.
\newblock In \emph{Artificial intelligence and statistics}, pp.\  1375--1383.
  PMLR, 2016.

\bibitem[Rifkin(2002)]{rifkin2002everything}
Rifkin, R.~M.
\newblock \emph{Everything old is new again: a fresh look at historical
  approaches in machine learning}.
\newblock PhD thesis, MaSSachuSettS InStitute of Technology, 2002.

\bibitem[Sangari \& Sethares(2015)Sangari and Sethares]{sangari2015convergence}
Sangari, A. and Sethares, W.
\newblock Convergence analysis of two loss functions in soft-max regression.
\newblock \emph{IEEE Transactions on Signal Processing}, 64\penalty0
  (5):\penalty0 1280--1288, 2015.

\bibitem[Sun(2019)]{sun2019optimization}
Sun, R.
\newblock Optimization for deep learning: theory and algorithms.
\newblock \emph{arXiv preprint arXiv:1912.08957}, 2019.

\bibitem[Tan \& Le(2019)Tan and Le]{tan2019efficientnet}
Tan, M. and Le, Q.
\newblock Efficientnet: Rethinking model scaling for convolutional neural
  networks.
\newblock In \emph{International conference on machine learning}, pp.\
  6105--6114. PMLR, 2019.

\bibitem[Towns et~al.(2014)Towns, Cockerill, Dahan, Foster, Gaither, Grimshaw,
  Hazlewood, Lathrop, Lifka, Peterson, et~al.]{towns2014xsede}
Towns, J., Cockerill, T., Dahan, M., Foster, I., Gaither, K., Grimshaw, A.,
  Hazlewood, V., Lathrop, S., Lifka, D., Peterson, G.~D., et~al.
\newblock Xsede: Accelerating scientific discovery computing in science \&
  engineering, 16 (5): 62--74, sep 2014.
\newblock \emph{URL https://doi. org/10.1109/mcse}, 128, 2014.

\bibitem[Vaswani et~al.(2017)Vaswani, Shazeer, Parmar, Uszkoreit, Jones, Gomez,
  Kaiser, and Polosukhin]{vaswani2017attention}
Vaswani, A., Shazeer, N., Parmar, N., Uszkoreit, J., Jones, L., Gomez, A.~N.,
  Kaiser, {\L}., and Polosukhin, I.
\newblock Attention is all you need.
\newblock \emph{Advances in neural information processing systems}, 30, 2017.

\bibitem[Watanabe et~al.(2018)Watanabe, Hori, Karita, Hayashi, Nishitoba, Unno,
  Soplin, Heymann, Wiesner, Chen, et~al.]{watanabe2018espnet}
Watanabe, S., Hori, T., Karita, S., Hayashi, T., Nishitoba, J., Unno, Y.,
  Soplin, N. E.~Y., Heymann, J., Wiesner, M., Chen, N., et~al.
\newblock Espnet: End-to-end speech processing toolkit.
\newblock \emph{arXiv preprint arXiv:1804.00015}, 2018.

\bibitem[Zagoruyko \& Komodakis(2016)Zagoruyko and
  Komodakis]{zagoruyko2016wide}
Zagoruyko, S. and Komodakis, N.
\newblock Wide residual networks.
\newblock \emph{arXiv preprint arXiv:1605.07146}, 2016.

\end{thebibliography}
